\documentclass[conference]{IEEEtran} 
\ifCLASSOPTIONcompsoc
  \usepackage[nocompress]{cite}
\else
  \usepackage{cite}
\fi
\ifCLASSINFOpdf
\else
\fi
\usepackage[colorlinks=true,
            linkcolor=red,
            urlcolor=blue,
            citecolor=blue]{hyperref}
\usepackage{subcaption}
\usepackage{cite}
\usepackage{amsmath,amssymb,amsfonts}
\usepackage[ruled,vlined,lined]{algorithm2e}
\usepackage{mathtools}
\usepackage{relsize}
\usepackage[scientific-notation=true]{siunitx}
\usepackage{cite}
\usepackage[noend]{algpseudocode}
\usepackage[utf8]{inputenc}
\usepackage{graphicx}
\usepackage{latexsym}
\usepackage{etoolbox}
\usepackage{textcomp}
\usepackage{xcolor}
\makeatletter
\usepackage[inline]{enumitem}
\def\BState{\State\hskip-\ALG@thistlm}
\makeatother

\AfterEndEnvironment{table}{\vskip-1ex}

\begin{document}

\title{Graph Convolution Neural Network For Weakly Supervised Abnormality Localization In Long Capsule Endoscopy Videos}

%
%

\markboth{IEEE Transactions on Medical Imaging}%
{Shell \MakeLowercase{\textit{et al.}}: Bare Demo of IEEEtran.cls for IEEE Journals}
%




\author{\IEEEauthorblockN{Sodiq Adewole\IEEEauthorrefmark{1},
Philip Fernandes \IEEEauthorrefmark{2},
James Jablonski \IEEEauthorrefmark{1}, 
Andrew Copland \IEEEauthorrefmark{2},
Michael Porter \IEEEauthorrefmark{1},\\
Sana Syed \IEEEauthorrefmark{2}, and
Donald Brown\IEEEauthorrefmark{1}\IEEEauthorrefmark{3}}

\IEEEauthorblockA{\IEEEauthorrefmark{1} Department of Systems and Information Engineering,
University of Virginia,
Charlottesville, VA, USA}

\IEEEauthorblockA{\IEEEauthorrefmark{1} Department of Pediatrics, School of Medicine,
University of Virginia,
Charlottesville, VA, USA}

\IEEEauthorblockA{\IEEEauthorrefmark{3} School of Data Science, 
University of Virginia,
Charlottesville, VA, USA}
}

\maketitle

\begin{abstract}
Temporal activity localization in long videos is an important problem. The cost of obtaining frame level label for long Wireless Capsule Endoscopy (WCE) videos is prohibitive. In this paper, we propose an end-to-end temporal abnormality localization for long WCE videos using only weak video level labels. Physicians use Capsule Endoscopy (CE) as a non-surgical and non-invasive method to examine the entire digestive tract in order to diagnose diseases or abnormalities. While CE has revolutionized traditional endoscopy procedures, a single CE examination could last up to 8 hours generating as much as 100,000 frames. Physicians must review the entire video, frame-by-frame, in order to identify the frames capturing relevant lesion or abnormality. This, sometimes could be as few as just a single frame. Given this very high level of redundancy, analysing long CE videos can be very tedious, time consuming and also error prone. This paper presents a novel multi-step method for an end-to-end localization of target frames capturing abnormalities of interest in the long video using only weak video labels. First we developed an automatic temporal segmentation using change point detection technique to temporally segment the video into uniform, homogeneous and identifiable segments. Then we employed Graph Convolutional Neural Network (GCNN) to learn a representation of each video segment. Using weak video segment labels, we trained our GCNN model to recognize each video segment as abnormal if it contains at least a single abnormal frame. Finally, leveraging the parameters of the trained GCNN model, we replaced the final layer of the network with a temporal pool layer to localize the relevant abnormal frames within each abnormal video segment. We experimented with multiple real patients' endoscopy videos and achieved an accuracy of 89.9\% on the graph classification task and a specificity of 97.5\% on the abnormal frames localization task.
\end{abstract}


\section{Introduction}
Gastrointestinal (GI) diseases are a source of substantial morbidity, mortality, and cost in the United States. In 2015, annual health care expenditures for gastrointestinal diseases totaled \$135.9 billion \cite{peery2019burden}. Endoscopy is the standard non-surgical procedure that allows physicians to examine the digestive tract to identify any disease, abnormalities or lesions present in the system. Traditional endoscopy procedures include the upper endoscopy, small bowel endoscopy and colonoscopy. In upper endoscopy, an endoscope is passed through the mouth and throat and into the esophagus, thereby allowing the physician to view the esophagus and stomach \cite{swain2003wireless}. The small-bowel endoscopy advances further and allows visibility into the upper part of the small intestine while colonoscopy involves passing endoscopes into the colon through the rectum to examine the colon. While these traditional methods are still very well in use today, the main limitation is their inability to provide visibility into significant part of the small bowel region. This is in addition to being uncomfortable, invasive and always requiring the physical presence of the physician or expert gastroenterologist.

\begin{figure}
    \centering
    \includegraphics[scale=0.2]{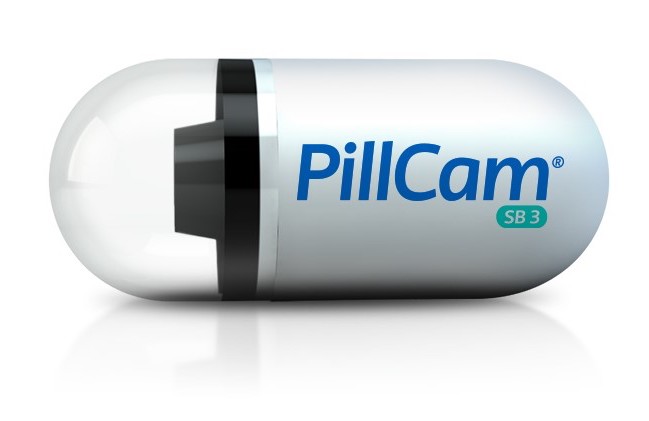}
    \caption{Capsule Camera}
    \label{fig:capsule}
\end{figure}

Wireless Capsule Endoscopy (WCE) \cite{iddan2000wireless} has revolutionized the traditional procedures by allowing for non-invasive visualization of the entire digestive tract including the entire small bowel region. During the WCE procedure, patients swallow a tiny capsule equipped with a camera (shown in fig. \ref{fig:capsule}) which captures images of the entire digestive tract at about 2 - 6 frames per second. The capsule is propelled down the digestive tract through peristaltic movement of the intestinal walls. The images are transmitted to an attached recorder and then transferred to a work station. The images are compiled as a video for review by an expert physician or gastroenterologist. Since traditional methods are limited in how much visibility they can provide, physicians are usually more interested in the images of the small bowel region when reviewing CE videos. While there could be as much as 50,000 images covering the small bowel region alone \cite{swain2003wireless}, it is possible for abnormality or lesion of interest to be present in as few as a single frame. However, physicians must review and analyze the entire video, frame-by-frame, in order to identify the abnormalities for diagnosis. This manual review process is very tedious, time consuming and also prone to error leading to increased risk of wrong diagnosis.

Automated analysis of videos encompasses tasks such as object detection, object recognition, tracking, action localization and general understanding of objects behavior in a video. Each of these task leverage a single or multiple information contained in the video for prediction. For example, object detection only require 2D image spatial data in order to identify the region of interest. Meanwhile object tracking will leverage both the spatial and temporal information contained in the video to perform the task. Due to the extreme difficulty of manually analysing video structured data by humans, various methods have been proposed in literature to automate some of the tasks across multiple domains \cite{shou2016temporal, perazzi2017learning, gao2017video}, including medical domain on WCE videos \cite{chen2017deep, tsuboi2020artificial, iobagiu2008colon}.

Activity localization or action detection \cite{yeung2016end, ghanem2017activitynet} in a video involves identifying the region where the activation score corresponding to the class of activity in the video is maximum. Activity localization in a short video has received significant attention among computer vision research community \cite{shou2016temporal, Zhao2017Apr, Gao2017May, Lin2017Oct, Shou2017Mar, Zhao2017Apr, Shou2016Jan, Soomro2015, Gkioxari2014Nov, Yeung2015Nov, Ma2016Jun, Buch2017Jul, Escorcia2016Sep, Wang2016Apr}. However, models such as structured segment network in \cite{Zhao2017Apr}, multi-stage CNN model in \cite{Shou2016Jan} and boundary regression in \cite{Gao2017May} requires frame-level labels to train. Obtaining frame annotation in medical domain, particularly for CE video data is very challenging. In order to develop a model that generalizes across multiple patients and diseases, the model would require large sample of each abnormality collected across multiple patients. Furthermore, while Deep Convolutional Neural Network (DCNN) based models have demonstrated improved performance on various image recognition \cite{deng2009imagenet, Simonyan2014Sep, Krizhevsky2014Apr} and video analysis tasks \cite{idrees2017thumos, cai2018weakly, chen2019weakly, li2021weakly, mahasseni2017unsupervised} including medical image analysis \cite{chen2017deep, rahim2020survey, miaou2009multi, sainju2014automated, yuan2015saliency}, they are notoriously sample inefficient requiring large samples per training class to optimized its parameters for ease of generalizability. The challenge of obtaining frame level label is further exacerbated, in the medical domain, when the expertise, time and effort required are not readily available. Despite the large volume of frames generated in a single CE examination, the high class imbalance, with significantly more normal frames than disease-containing frames, limits the feasibility of training a fully supervised DCNN model that generalizes across multiple abnormalities and also patients.

Prior works on CE video have mostly focused on single or multiple lesion detection on each individual and independent frames in the video \cite{nawarathna2014abnormal, rahim2020survey, li2009computer, miaou2009multi, mewes2012semantic, kodogiannis2008neuro, zhao2010abnormality, sainju2014automated, van2009capsule, mamonov2014automated, hwang2010polyp, yuan2015improved, tsuboi2020artificial, pogorelov2018deep, deeba2018saliency, gao2020deep, yuan2015saliency, adewole2020deep, chen2009developing, mackiewicz2008wireless, vu2009detection, chen2012review, sivakumar2019novel, eliakim2013video, adewole2021lesion2vec, iobagiu2008colon}. Despite the extreme difficulty of obtaining frame-level label for CE video frames, little to no attention has been made towards leveraging temporal or topological relationship between the frames to develop a more robust system. To the best of our knowledge, no prior work has addressed the task of temporal abnormality localization within a sequence of CE video frames. We believe that analysis of video data requires leveraging the spatial, temporal and topological relationship between the frames to achieve a system that can be deployed in real clinical environment to aid physicians in their diagnosis. The novelty of the work proposed in this paper is in three (3) folds; Firstly, we leverage the spatial, temporal and topological relationship between the frames to develop a model to localize abnormal regions containing the disease or abnormality of interest in a full CE video. Secondly, our model uses only weak video level labels for this task, thereby obviating the need for an expert provided frame-level annotation, which is often very challenging. Thirdly, we employed Graph Convolutional Neural Network (GCNN) model, based on the GraphSage architecture \cite{Hamilton2017Jun}. This allows us to learn a robust representation of CE videos both transductively and inductively by leveraging the message passing architecture and neighborhood information aggregation.

Different techniques have been proposed for lesion segmentation within a 2-D CE video frame \cite{zhao2010abnormality, hwang2010polyp, mackiewicz2008wireless}. Similar to the high cost of obtaining pixel-level label for image segmentation, obtaining frame-level labels for CE videos is not an easy task. First, annotating individual frame is much more tedious than the normal CE video review process. Secondly, challenging conditions such as poor illumination and camera instability due to peristaltic motion of the bowel impacts the quality of frames generated in the video, leading, sometimes, to noisy and unreliable expert-provided frame-level annotation. To the best of our knowledge, this is the first work on abnormality localization in a sequence of CE video frames using weak video level labels. The model proposed in this paper, addresses several issues in CE video analysis where our weakly supervised model requires no frame-level annotation from medical experts. In addition, by using graph-based model, we learn a more robust representation of the video through message passing and information aggregation.

Given weak labels for each video segment, we train a weakly supervised GCNN model on aggregate frame features and classify each video segment. During testing, we applied a adaptive temporal pool layer on the GCNN model to generate frames' activation score corresponding to the video class activation map over the sequence of frames. The adaptive temporal pool layer ranks the frames within each segment based on the significance to identifying the segment as abnormal. This significance of this framework is in minimizing experts' review time on CE videos by generating frames relevant to abnormality of interest for review by the expert physician or gastroenterologist without the need for frame level labels.

Long videos typically differs from short videos based on the duration and also the number of actions contained in the sequence. Since short videos usually contain one object or activity of interest, activity localization within a short video involve detecting a single high energy region in the sequence. Meanwhile, long videos pose additional challenge with multiple energy activation regions requiring temporal segmentation before localization. In addition to the novelty of the work previously mention, with the end-to-end system for long videos proposed in this paper, we are able to generalize the concept activity localization to long videos with multiple activities within the sequence. Without requiring manual partitioning of the video into fixed frame length. Localizing action in short videos involves a temporal search for a single class activation map within the sequence while long videos with multiple activities will have multiple actions withing the sequence.

The rest of the paper is structured as follows: Section \ref{sec:related_work} covers related work on GCNN model, weakly supervised learning and prior works on analysis of CE video data. In section \ref{sec:method}, we cover the main idea of the framework proposed in this paper including detailed description of each component of our multi-step localization model. Section \ref{sec:experiments} describes our experimentation procedure, including detailed description of our dataset, evaluation metrics and implementation procedures. Section \ref{sec:results} contains the results of our experiments on multiple patients CE video data including the result our our ablation study where we varied the $k$ hyper-parameter in our sample frame localization. In section \ref{sec:conclusion}, we summarized the results of the framework developed in this paper, our contribution, limitations and our next steps for future works.

\subsection{Contributions:} (1) A novel temporal abnormality localization model for long capsule endoscopy video; (2) Using Graph Convolution Neural Network (GCNN) trained only weak video level annotation, we employed information present in all the nodes of the graph to learn a better representation of each video in our data; (3) We conducted ablation study using different configuration of the GraphSage-GCNN model to understand the effect of each component on the performance of the network; (4) We reported performance of the model on the localization task using a range of hyper-parameters thresholds for the final temporal pool layer of the network.
\section{Related Work}\label{sec:related_work}
In this section, we discuss the prior works on analysis of CE video as well as techniques that have been developed for various disease and abnormality detection. Our review covers work on GCNN in other domains and different formulations and solutions to weakly supervised learning tasks.
\subsection{Abnormality Detection in Capsule Endoscopy Videos}
Analysis of CE videos encompasses tasks such as disease or abnormality detection, quantifying severity of identified diseases, localizing identified abnormalities, and decision making on appropriate intervention by the physician. Prior works on automating review and analysis of CE videos can broadly be categorized into three (3) - 1) detection of specific disease or lesion such as bleeding in \cite{sainju2014automated}, polyp  \cite{mamonov2014automated}, ulcer \cite{yuan2015saliency}, and angioectasia \cite{tsuboi2020artificial, pogorelov2018deep}; 2) abnormal or outlier frame detection where frames with abnormalities are consider outliers \cite{gao2020deep, zhao2010abnormality};  and 3) models aimed at minimizing experts review time on CE video - video summarization. Here key frames capturing abnormalities are selected as representative frames from the entire video \cite{iakovidis2010reduction, emam2015adaptive, mehmood2014video, mohammed2017sparse, ismail2013endoscopy, chen2016wireless}. While obtaining frame level label for CE videos is very difficult, little to no attention has been paid to models that will leverage the relationship between the frames to mitigate this challenge. To the best of our knowledge, no prior work has considered temporal abnormality localization on CE video data. The work proposed in this paper aligns with the concept of video summarization where, by leveraging the temporal and topological relationship between the video frames, we localize the abnormality to a more narrow temporal region. This allows us to select representative samples within each abnormal region as a video summary for the experts. In addition, our model does not require any frame level label to identify the abnormal regions and localize abnormal frames in the video.

\subsection{Graph Convolutional Neural Network (GCNN)}
Following the work in \cite{kipf2016semi}, GCNN continues to gain increased popularity among deep learning and machine learning researchers\cite{rasoul2021feature}. GCNN extends techniques such as \textit{Recursive Neural Networks (RNN)} \cite{Hagenbuchner2003May, Sperduti1997May} and \textit{Markov Chains} \cite{Tsoi2003Jan, Kleinberg1999Sep} while leveraging the powerful representation power of neural networks on graph structured data. Traditional deep learning models are well developed for spatial (CNN) \cite{adewole2021lesion2vec, sali2020hierarchical} and sequential (RNN) data \cite{adewole2020dialogue} with little contribution on graph structured data. CNNs are used to learn representation on 2D spatial image data while RNNs learns to encode and represent sequential data. While CNNs and RNNs based models \cite{chen2016wireless} can automatically learn the internal encoding of the graph structured data, SVM's internal representation needs to be user designed. Meanwhile, many natural interactions between objects can be represented as a graph with the relationship between the objects captured in the edges between the nodes of the graph. Graph Neural Networks (GNN) models are robust and generic enough to also accommodate spatial and sequence data \cite{sharma2019data, adewole2020dialogue} by specifying the nature of the edge and node relationships. 

Main operations on graph network are filtering, activation and pooling. Similar to regular convolution, Graph Convolution Network (GCNN) combines the benefit of spatial and spectral based filtering operations \cite{kipf2016semi} in addition to non-linear transformation of the input features to achieve a robust representation of the graph structured data. GCNN represents features as nodes in the graph and wide range of relationships, from simple similarity (e.g. cosine similarity) to long- short term memory (LSTM) can be modelled to capture the relationship between the nodes as weighted edges. Graph filtering uses neighborhood aggregation from the previous layer to determine the representation of each node in subsequent layer \cite{Hamilton2017Jun}. \cite{Hamilton2017Jun} proposed GraphSage to leverage both inductive and transductive learning capability of GNN. For each layer of the network, the model aggregates the representation for each node in the graph based neighborhood sampling from surrounding nodes. Graph Attention Network (GAT) \cite{velivckovic2017graph} was proposed to improve the neighborhood aggregation by ranking the neighboring nodes using an attention layer to generate better representation.

\subsection{Weakly Supervised Localization}
State-of-the-art methods address the problem of temporal action localization in long videos by applying RNN based action classifiers on sliding windows \cite{yeung2016end, ghanem2017activitynet} for action detection in a video sequence. Methods such as structured segment network in \cite{Zhao2017Apr}, multi-stage CNN model in \cite{Shou2016Jan} and boundary regression in \cite{Gao2017May} are some of the approaches to action detection in a sequence of video frames. However, these techniques require frame level annotation which is a very difficult to collect, particularly in medical domain \cite{malik2021ten}. In order to mitigate this challenge, weakly supervised methods using global video level labels for activity localization has recently been gaining traction among researchers \cite{nguyen2018weakly, Nguyen2017Dec, Paul2018Jul, bae2020rethinking, Rashid_2020_WACV}. In \cite{Nguyen2017Dec} Nguyen et al., proposed sparse temporal pooling network for action localization in an untrimmed video. Using video-level class labels, their model predicts temporal intervals of human actions in a video. In \cite{Paul2018Jul} the authors proposed the Weakly supervised Temporal Activity Localization and Classification (W-TALC) framework using only video-level labels. They used two sub-networks - a two-stream based feature extractor network and a weakly-supervised module - trained by optimizing two complimentary loss functions. The model learns to classify the videos and also localize the region of the action within the video. Class Activation Mapping (CAM) was introduced in \cite{bae2020rethinking} for weakly supervised action localization in an untrimmed video. Similarly, \cite{Gao2017May} proposed a cascaded boundary regression method for temporal action localization.

In \cite{Lin2017Oct}, Lin et al. proposed a single shot technique for temporal action detection in a video. Their model based on 1D temporal convolutional layers, skips the proposal generation step in detection by classification framework, to directly detect action instances in untrimmed videos. \cite{Shou2017Mar} used convolution de-convolution network to precisely localize action in untrimmed videos. The work in \cite{Rochan_2015_CVPR} is focused on weakly supervised localization of novel objects using the objects' appearance transfer framework. Another unique attempt at action localization was proposed in \cite{Shou_2018_ECCV}, where the authors temporally localized action in untrimmed videos using (Auto-loc). UntrimmedNets was proposed in \cite{Wang2017Mar} for temporal action recognition and detection.

While our proposed framework is motivated by \cite{nguyen2018weakly, Rashid_2020_WACV, Hamilton2017Jun}, our model combines more effective GraphSage representation network with a final attention layer in the classification model. As against just simple temporal attention model used in \cite{Nguyen2017Dec}, our GCNN representation is able to leverage the neighborhood information for more effective representation of each member node in the graph. However, we adapted the temporal pool layer based on \cite{nguyen2018weakly} for the abnormal frame localization during inference. GCNN localization framework was considered in \cite{Rashid_2020_WACV}, our model is different in that the aim of our localization task is to select sparse representative frames in each video segment as against using similarity between time segments to determine the temporal boundaries \cite{Rashid_2020_WACV}. Secondly, this paper addresses the problem of temporal abnormality localization in long CE videos which is collected under more unstable and challenging digestive tract environment than most open dataset. Lastly the peculiarity of this work as against other prior works on activity localization is that abnormal regions in CE videos are not usually contiguous, making frameworks developed temporal segment boundary detection ineffective. Our model, therefore aims to select sparse non-contiguous representative frames within each video segment by applying a temporal pool layer over the final GCNN activation layer. To the best of our knowledge, this is the first work using temporal information to localize abnormal frames in CE video data. 

\section{Methodology}\label{sec:method}
\subsection{Overview of Framework}
Weakly supervised temporal abnormality localization is an extension of weakly supervised object segmentation task on 2-dimensional image data \cite{zhao2010abnormality, hwang2010polyp, mackiewicz2008wireless, Jie2017Apr, bilen2016weakly, Kantorov2016Sep}. In this section, we describe our framework on the multi-step abnormality localization on CE video data. We employed GraphSage model as the base GCNN model. First we temporally segment the long video into uniform homogeneous and identifiable short video segment. Then, considering each video segment as a graph, the frames in the video represents the nodes while the relationship between the frames is represented as the edges of the graph. For our framework, we experimented with a range similarity measures such as cosine similarity, correlation and euclidean distance between the nodes of the graph to capture the topological relationship between the nodes in the edge weights. By using GCNN, we explicitly capture the topological relationships between the frames in the videos which is considered during training and testing of the network. For each layer of the graph, each node (frame) feature is transformed to a weighted average of the neighboring feature of the previous layer. In final layer of the network, we applied an attention mechanism to attend over the most informative frame in the sequence. Finally, we aggregated the weighted features after attention using an aggregator function. We also experimented with aggregator functions ranging from simple summation, mean and pool aggregation functions to a more complex LSTM aggregator to compare performance. The final layer is a multiple-instance classifier which is a fully connected layer that learns a multi-instance classification  \cite{carbonneau2018multiple} of the video. During inference, we perform the localization task by replacing the fully connected multi-instance classifier with a temporal pool layer on top of the last attention layer. This allows us to extract the frames with highest class activation score over the temporal sequence of frames in the region. Figure \ref{fig:loc_model} shows the pipeline of our proposed architecture.
\subsection{Feature Extraction}
Representation learning for medical images to capture abnormality of interest is a very complex problem. The nature of different categories of abnormalities in CE video further exacerbate this problem. First, the challenging environment in which the video frames are captured such as poor illumination and unstable movement of the capsule camera in the digestive tract lead to poor quality frames being generated. Secondly, abnormal lesion such as angioectasia, ulcer and polyp varies significantly with complex geometry and coloration, making detecting them in the video frames very difficult. To mitigate this problem, we compared multiple feature extraction approaches on the CE video frames, and adopted the VGG-19 \cite{Simonyan2014Sep} network which is in line with the result in \cite{adewole2020deep}. VGG-19 better captures and also learns better representation the different abnormalities present in the frames. We employed a pretrained VGG-19 network trained on large ImageNet dataseet \cite{simonyan2014very} and then fine-tuned on five (5) of our CE videos. 

CE videos data have high imbalance and high redundancy structure with far more normal frames than any abnormal category. While fine-tuning the model, we applied weighted oversampling on the classes by placing more weights on the minority class to create a balanced exposure of the model to the different classes. We also applied different augmentations such as random rotation, random horizontal and vertical flip to simulate the real movement of the capsule camera in the digestive tract. While this does not change the distribution of our data, we were able to generate additional samples using these augmentation techniques. Finally, we obtained a 4096-dimensional feature vector per frame from the pool-5 layer of the network after training. Each unique patient's video is represented by feature volume of $\text{T x d}_{in}$ where $T$ is the length (number of frames) of the video while $d_{in}$ is the dimension of each frame extracted from the pretrained VGG-network.

\subsection{Temporal Segmentation}
Using concept from time series change point detection \cite{ducre2003comparison, reeves2007review, rybach2009audio, chowdhury2012bayesian, chib1998estimation, cho2015multiple} and video shots boundary detection \cite{smeaton2010video, smeaton2010video, lienhart1998comparison, hanjalic2002shot, boreczky1996comparison}, we temporally partition the long video feature matrix into uniform homogeneous segment by detecting points at which the statistical properties of the sequence of features changes. 

\begin{figure}
     \centering
     \begin{subfigure}[b]{0.1\textwidth}
         \centering
         \includegraphics[trim=32 32 32 32,clip,scale=0.04]{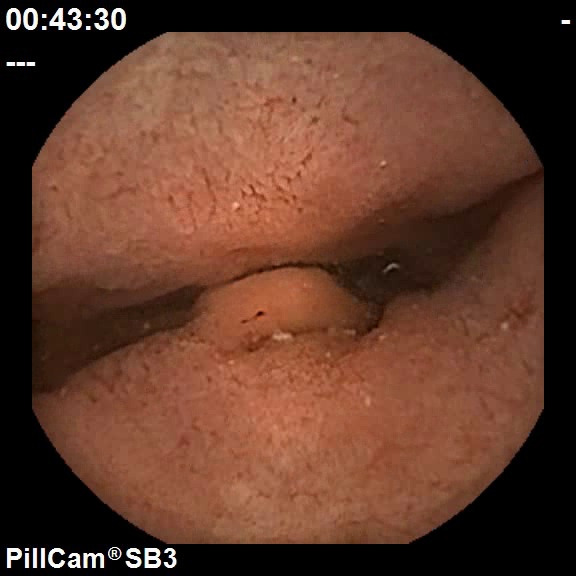}
     \end{subfigure}
     \hspace{-1.35cm}
      \begin{subfigure}[b]{0.1\textwidth}
         \centering
         \includegraphics[trim=32 32 32 32,clip,scale=0.04]{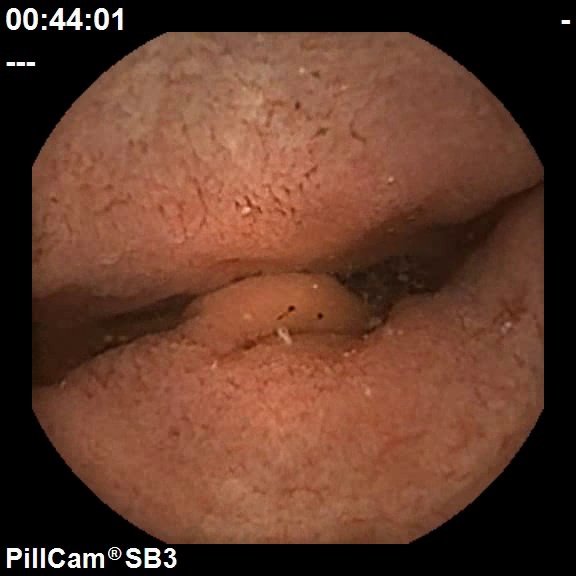}
     \end{subfigure}
     \hspace{-1.35cm}
      \begin{subfigure}[b]{0.1\textwidth}
         \centering
         \includegraphics[trim=32 32 32 32,clip,scale=0.04]{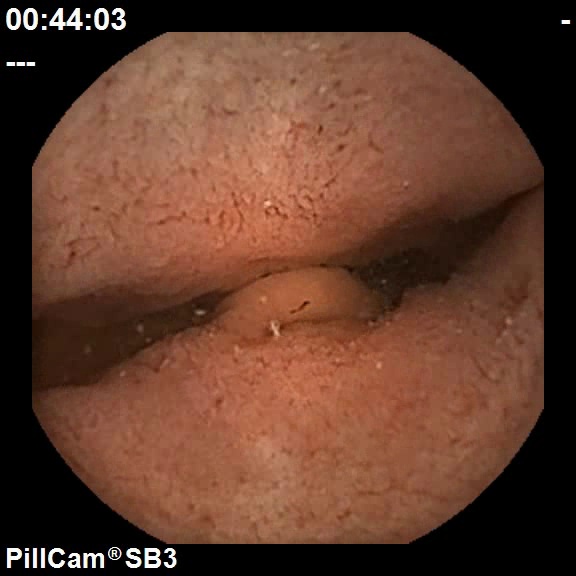}
     \end{subfigure}
    \hspace{-1.35cm}
      \begin{subfigure}[b]{0.1\textwidth}
         \centering
         \includegraphics[trim=32 32 32 32,clip,scale=0.04]{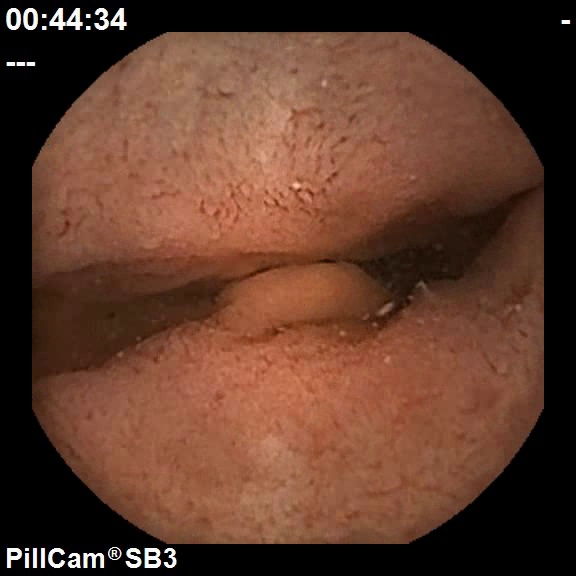}
     \end{subfigure}
    \hspace{-1.35cm}
      \begin{subfigure}[b]{0.1\textwidth}
         \centering
         \includegraphics[trim=32 32 32 32,clip,scale=0.04]{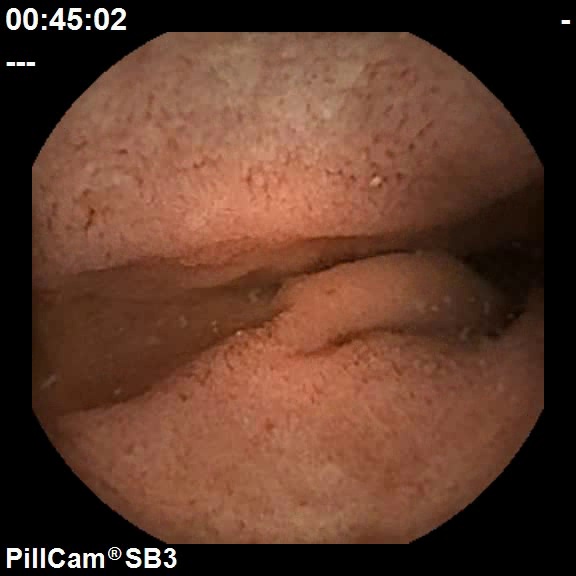}
     \end{subfigure}
    \hspace{-.750cm}
   \rule{1pt}{23pt}
    \hspace{-.750cm}
      \begin{subfigure}[b]{0.1\textwidth}
         \centering
         \includegraphics[trim=32 32 32 32,clip,scale=0.04]{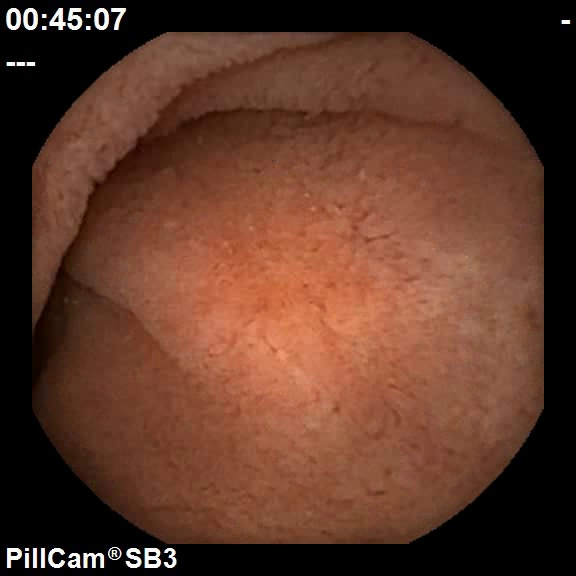}
     \end{subfigure}
    \hspace{-1.35cm}
      \begin{subfigure}[b]{0.1\textwidth}
         \centering
         \includegraphics[trim=32 32 32 32,clip,scale=0.04]{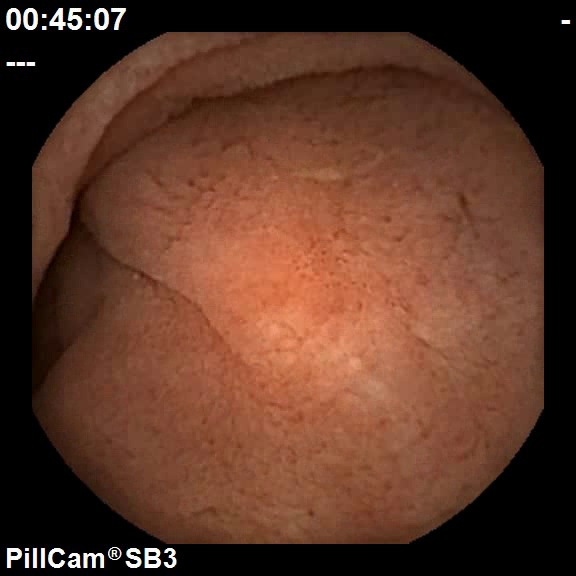}
     \end{subfigure}
    \hspace{-1.35cm}
      \begin{subfigure}[b]{0.1\textwidth}
         \centering
         \includegraphics[trim=32 32 32 32,clip,scale=0.04]{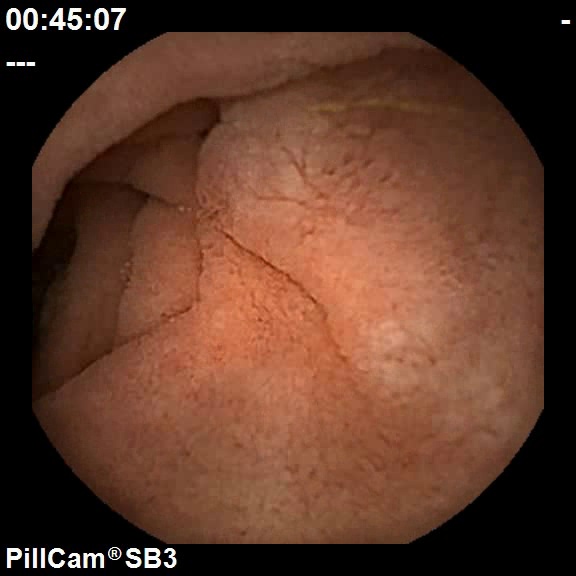}
     \end{subfigure}
    \hspace{-1.35cm}
      \begin{subfigure}[b]{0.1\textwidth}
         \centering
         \includegraphics[trim=32 32 32 32,clip,scale=0.04]{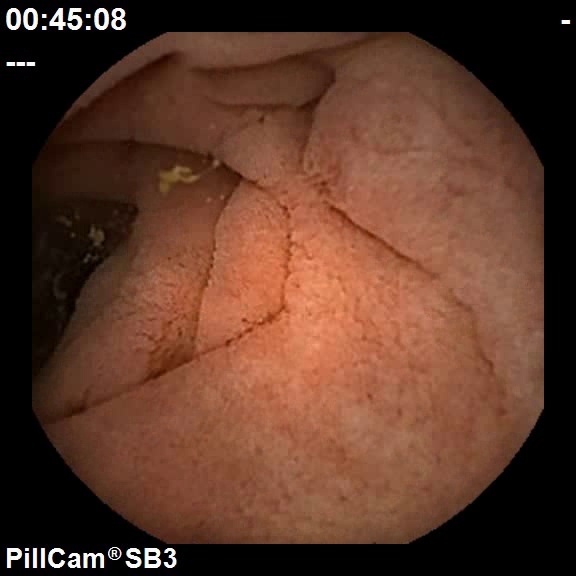}
     \end{subfigure}
    \hspace{-.750cm}
   \rule{1pt}{23pt}
    \hspace{-.750cm}
      \begin{subfigure}[b]{0.1\textwidth}
         \centering
         \includegraphics[trim=32 32 32 32,clip,scale=0.04]{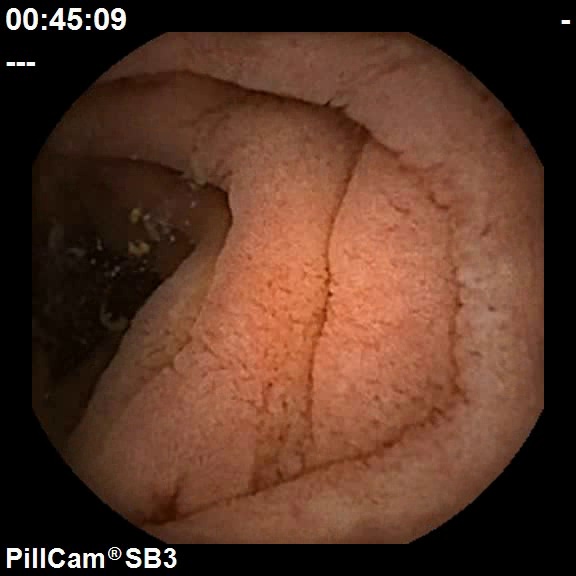}
     \end{subfigure}
     \hspace{-1.35cm}
      \begin{subfigure}[b]{0.1\textwidth}
         \centering
         \includegraphics[trim=32 32 32 32,clip,scale=0.04]{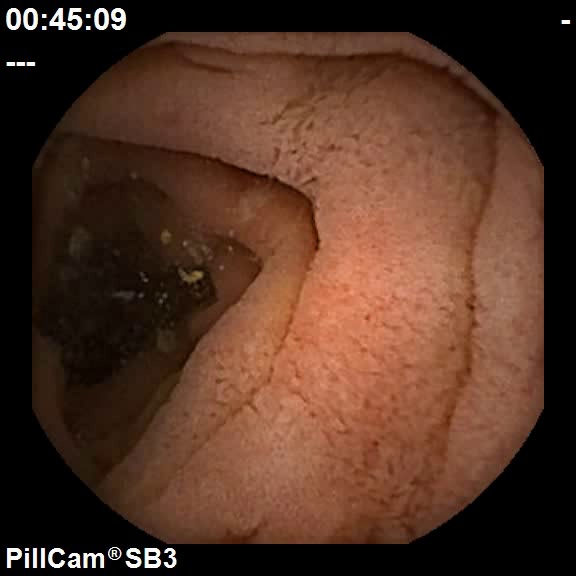}
     \end{subfigure}
     \hspace{-1.35cm}
      \begin{subfigure}[b]{0.1\textwidth}
         \centering
         \includegraphics[trim=32 32 32 32,clip,scale=0.04]{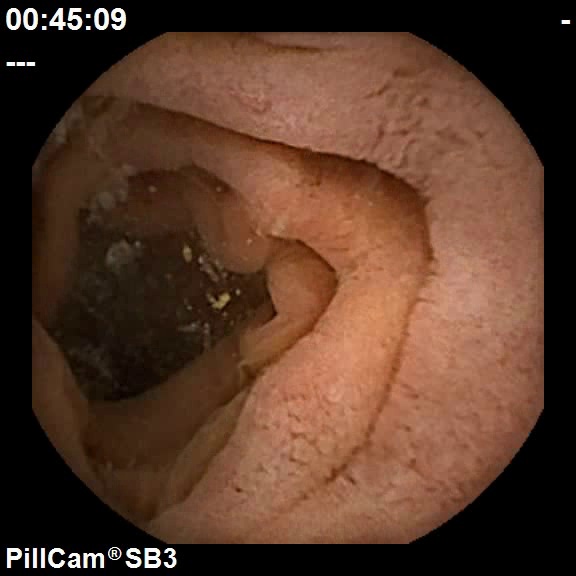}
     \end{subfigure}
    \hspace{-1.35cm}
      \begin{subfigure}[b]{0.1\textwidth}
         \centering
         \includegraphics[trim=32 32 32 32,clip,scale=0.04]{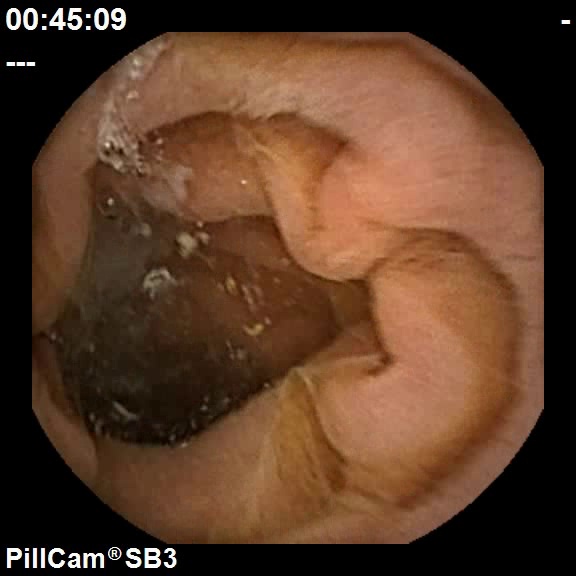}
     \end{subfigure}
    \hspace{-1.35cm}
      \begin{subfigure}[b]{0.1\textwidth}
         \centering
         \includegraphics[trim=32 32 32 32,clip,scale=0.04]{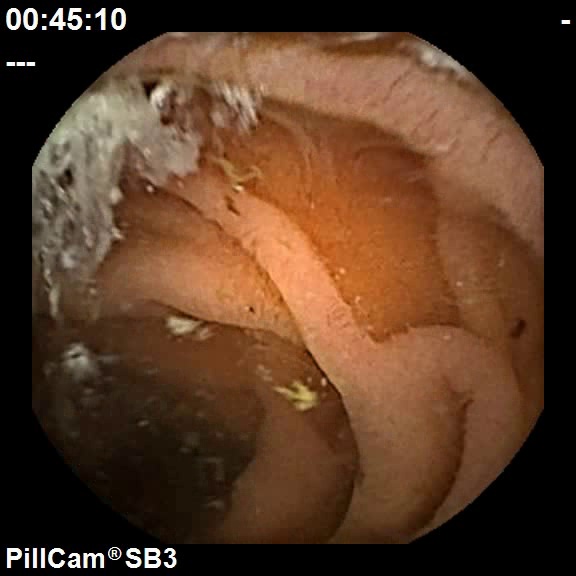}
     \end{subfigure}
    \hspace{-1.35cm}
      \begin{subfigure}[b]{0.1\textwidth}
         \centering
         \includegraphics[trim=32 32 32 32,clip,scale=0.04]{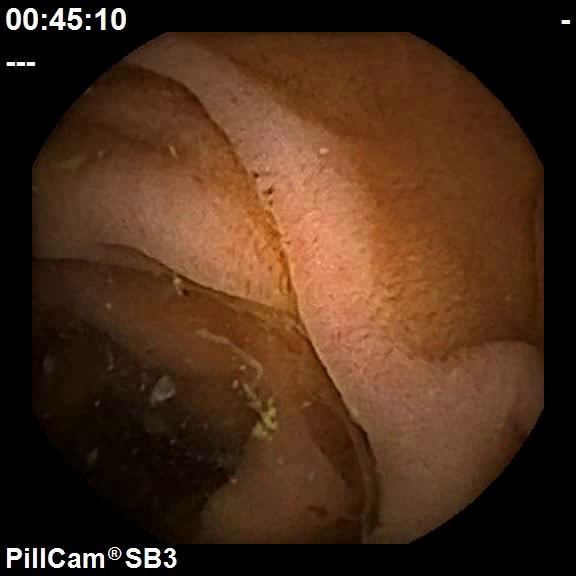}
     \end{subfigure}
    \hspace{-1.35cm}
      \begin{subfigure}[b]{0.1\textwidth}
         \centering
         \includegraphics[trim=32 32 32 32,clip,scale=0.04]{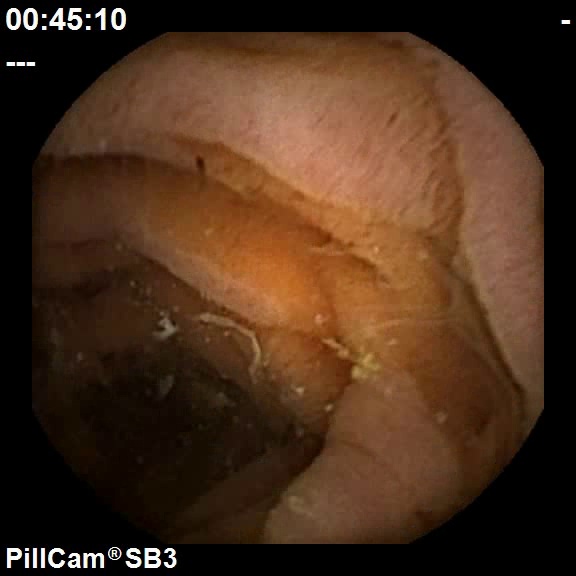}
     \end{subfigure}
    \hspace{-1.35cm}
      \begin{subfigure}[b]{0.1\textwidth}
         \centering
         \includegraphics[trim=32 32 32 32,clip,scale=0.04]{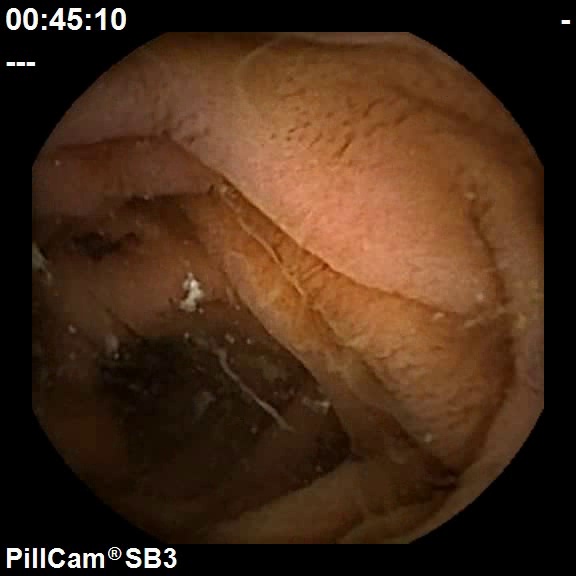}
     \end{subfigure}
    \hspace{-1.35cm}
      \begin{subfigure}[b]{0.1\textwidth}
         \centering
         \includegraphics[trim=32 32 32 32,clip,scale=0.04]{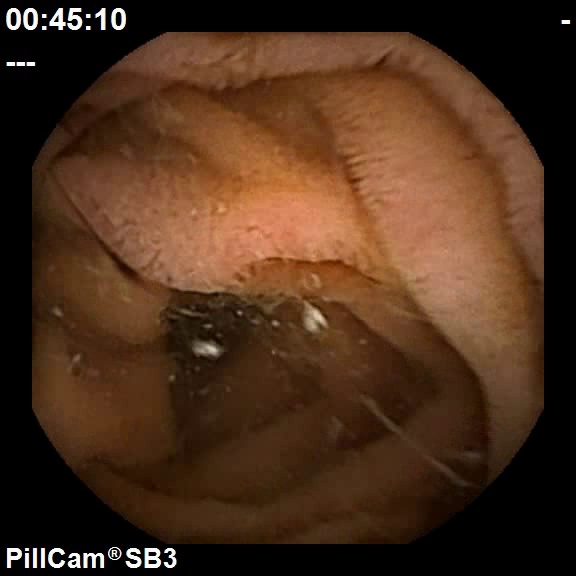}
     \end{subfigure}    
     \hspace{-1.35cm}
      \begin{subfigure}[b]{0.1\textwidth}
         \centering
         \includegraphics[trim=32 32 32 32,clip,scale=0.04]{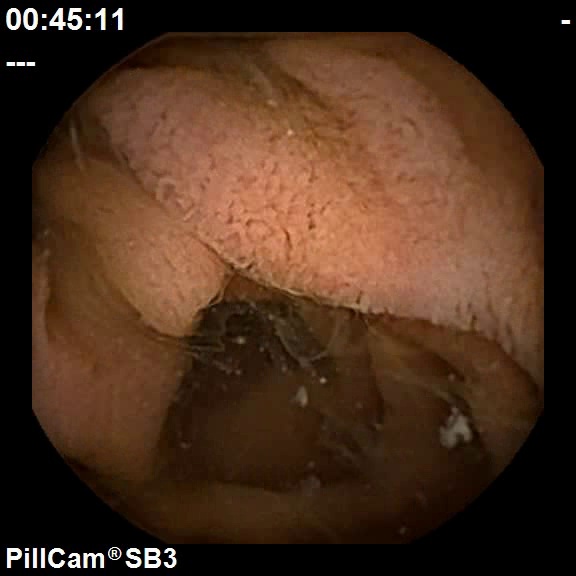}
     \end{subfigure}
    \hspace{-1.35cm}
      \begin{subfigure}[b]{0.1\textwidth}
         \centering
         \includegraphics[trim=32 32 32 32,clip,scale=0.04]{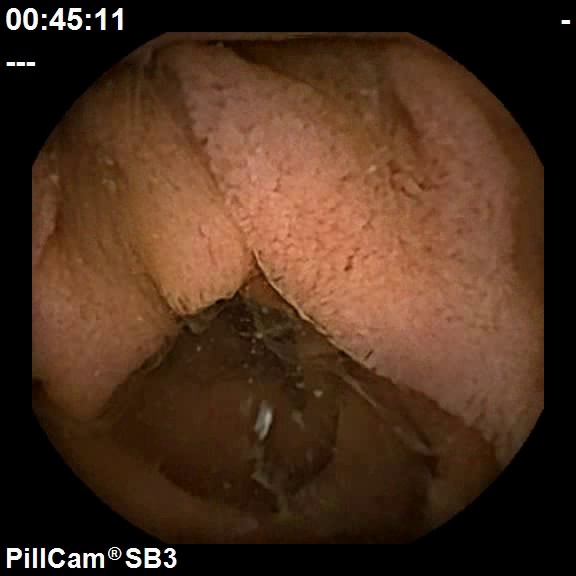}
     \end{subfigure}
     \hspace{-1.35cm}
    \begin{subfigure}[b]{0.1\textwidth}
         \centering
         \includegraphics[trim=32 32 32 32,clip,scale=0.04]{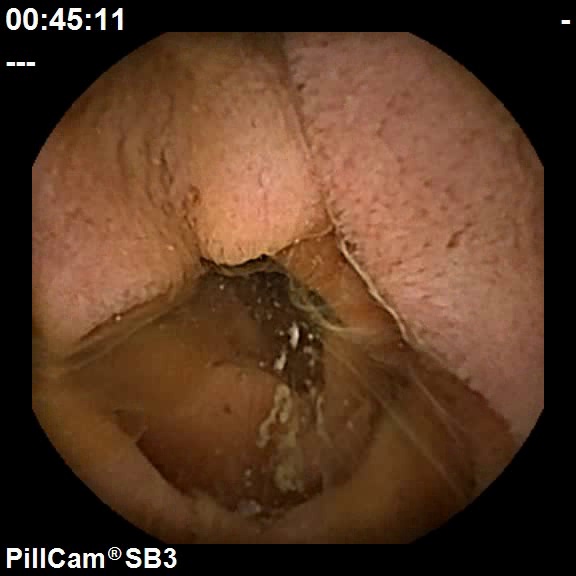}
     \end{subfigure}
     \hspace{-1.35cm}
      \begin{subfigure}[b]{0.1\textwidth}
         \centering
         \includegraphics[trim=32 32 32 32,clip,scale=0.04]{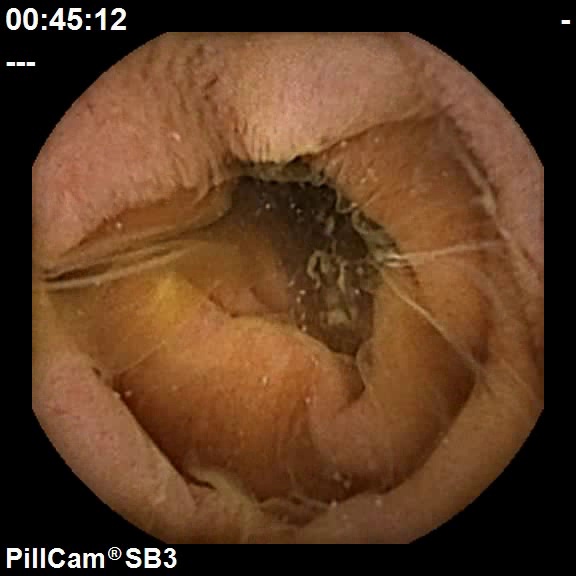}
     \end{subfigure}
     \hspace{-1.35cm}
      \begin{subfigure}[b]{0.1\textwidth}
         \centering
         \includegraphics[trim=32 32 32 32,clip,scale=0.04]{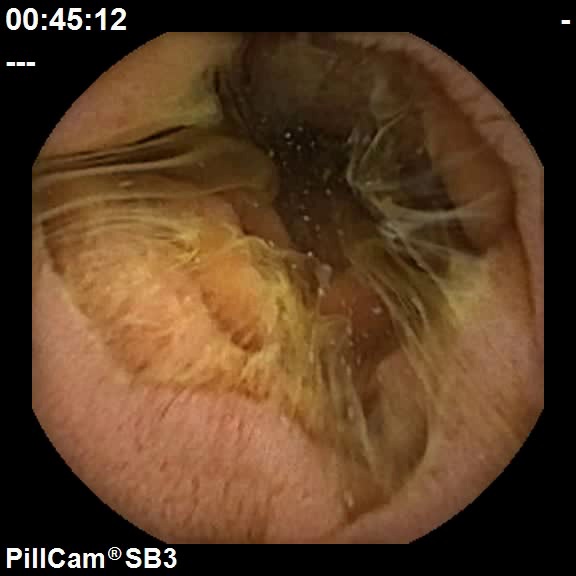}
     \end{subfigure}
    \hspace{-1.35cm}
      \begin{subfigure}[b]{0.1\textwidth}
         \centering
         \includegraphics[trim=32 32 32 32,clip,scale=0.04]{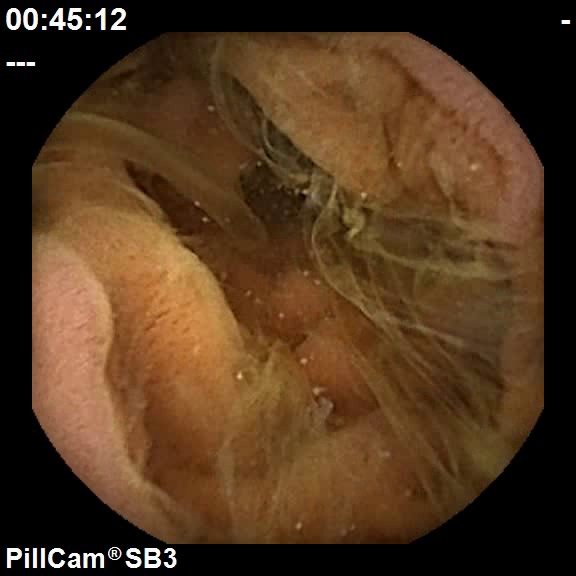}
     \end{subfigure}
    \hspace{-1.35cm}
      \begin{subfigure}[b]{0.1\textwidth}
         \centering
         \includegraphics[trim=32 32 32 32,clip,scale=0.04]{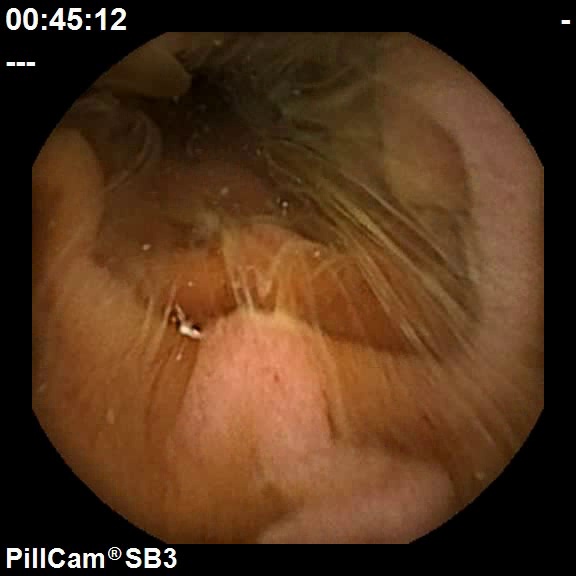}
     \end{subfigure}
    \hspace{-.750cm}
   \rule{1pt}{23pt}
    \hspace{-.750cm}
      \begin{subfigure}[b]{0.1\textwidth}
         \centering
         \includegraphics[trim=32 32 32 32,clip,scale=0.04]{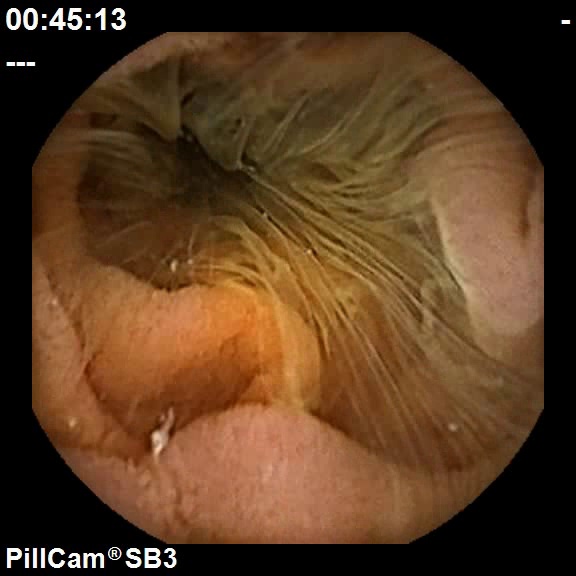}
     \end{subfigure}
    \hspace{-1.35cm}
      \begin{subfigure}[b]{0.1\textwidth}
         \centering
         \includegraphics[trim=32 32 32 32,clip,scale=0.04]{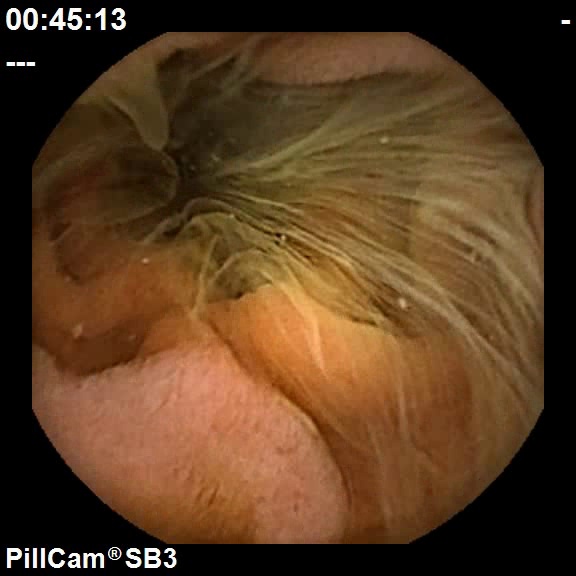}
     \end{subfigure}
    \hspace{-1.35cm}
      \begin{subfigure}[b]{0.1\textwidth}
         \centering
         \includegraphics[trim=32 32 32 32,clip,scale=0.04]{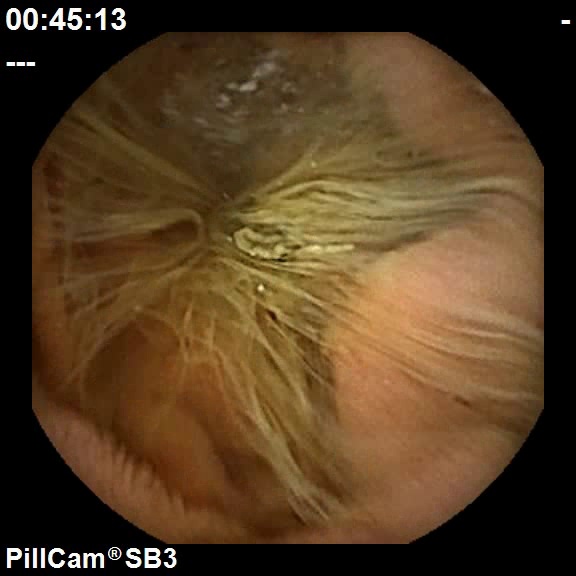}
     \end{subfigure}
    \hspace{-.750cm}
   \rule{1pt}{23pt}
    \hspace{-.750cm}
      \begin{subfigure}[b]{0.1\textwidth}
         \centering
         \includegraphics[trim=32 32 32 32,clip,scale=0.04]{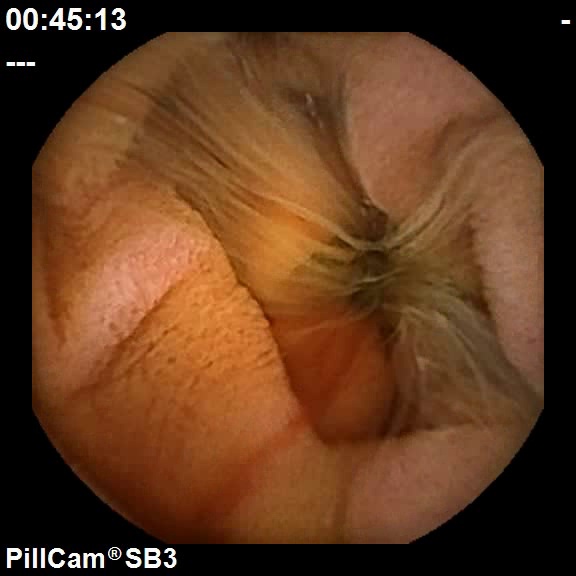}
     \end{subfigure}
     \hspace{-1.35cm}
      \begin{subfigure}[b]{0.1\textwidth}
         \centering
         \includegraphics[trim=32 32 32 32,clip,scale=0.04]{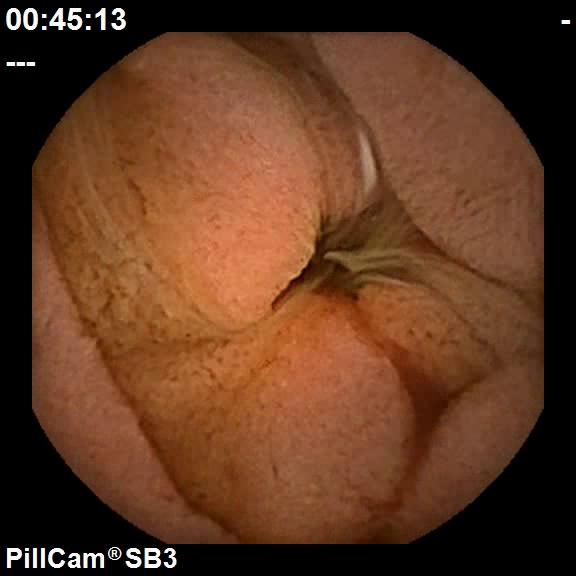}
     \end{subfigure}
     \hspace{-1.35cm}
      \begin{subfigure}[b]{0.1\textwidth}
         \centering
         \includegraphics[trim=32 32 32 32,clip,scale=0.04]{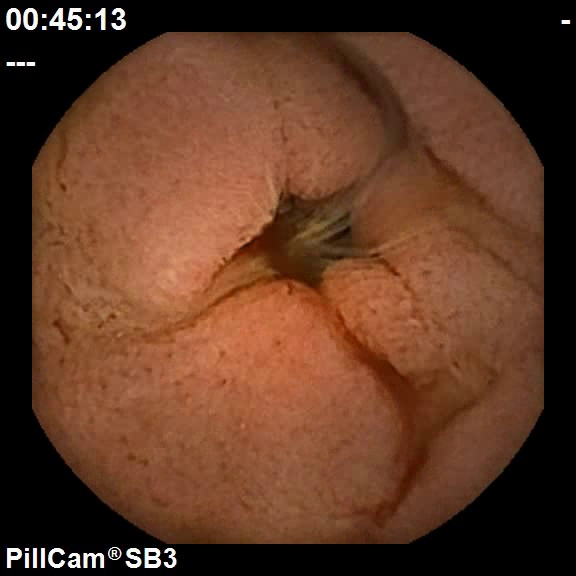}
     \end{subfigure}
     \hspace{-1.35cm}
      \begin{subfigure}[b]{0.1\textwidth}
         \centering
         \includegraphics[trim=32 32 32 32,clip,scale=0.04]{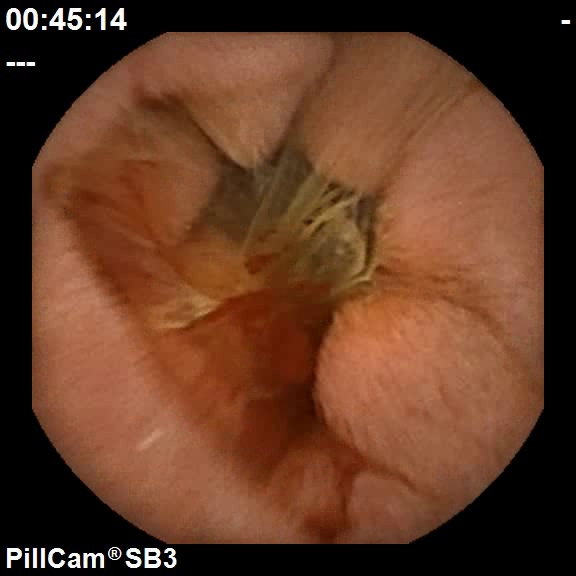}
     \end{subfigure}
    \hspace{-.750cm}
   \rule{1pt}{23pt}
    \hspace{-.750cm}
      \begin{subfigure}[b]{0.1\textwidth}
         \centering
         \includegraphics[trim=32 32 32 32,clip,scale=0.04]{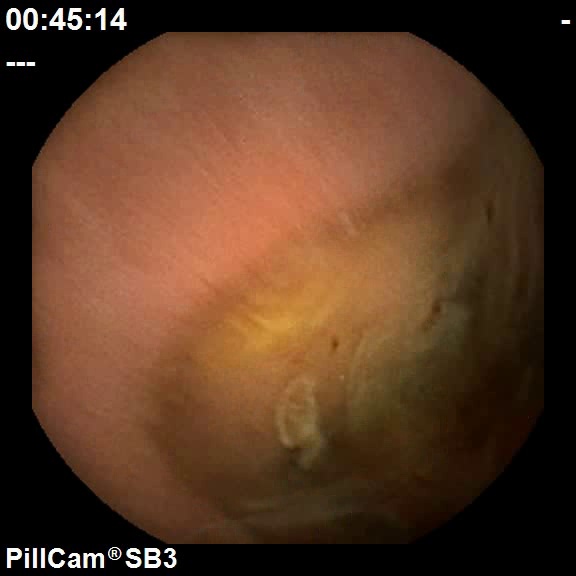}
     \end{subfigure}
    \hspace{-1.35cm}
      \begin{subfigure}[b]{0.1\textwidth}
         \centering
         \includegraphics[trim=32 32 32 32,clip,scale=0.04]{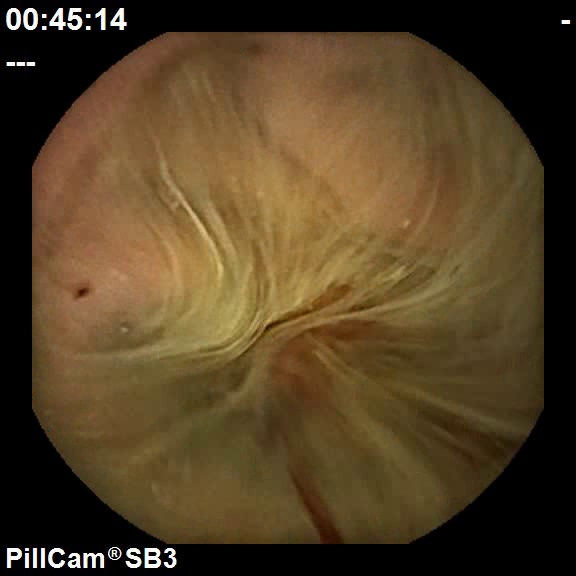}
     \end{subfigure}
    \hspace{-1.35cm}
      \begin{subfigure}[b]{0.1\textwidth}
         \centering
         \includegraphics[trim=32 32 32 32,clip,scale=0.04]{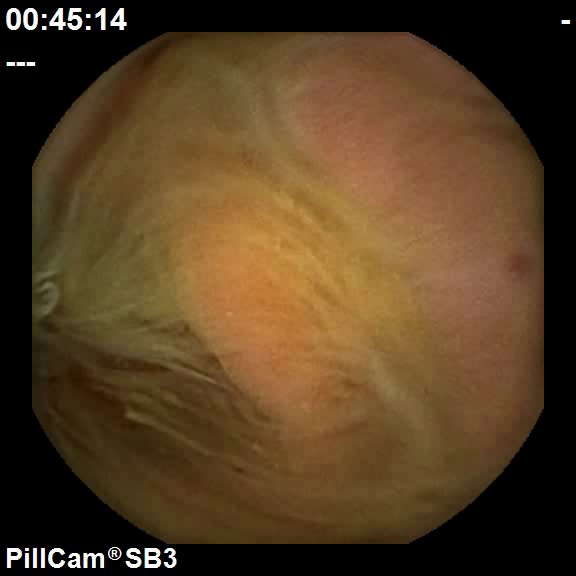}
     \end{subfigure}
    \hspace{-1.35cm}
      \begin{subfigure}[b]{0.1\textwidth}
         \centering
         \includegraphics[trim=32 32 32 32,clip,scale=0.04]{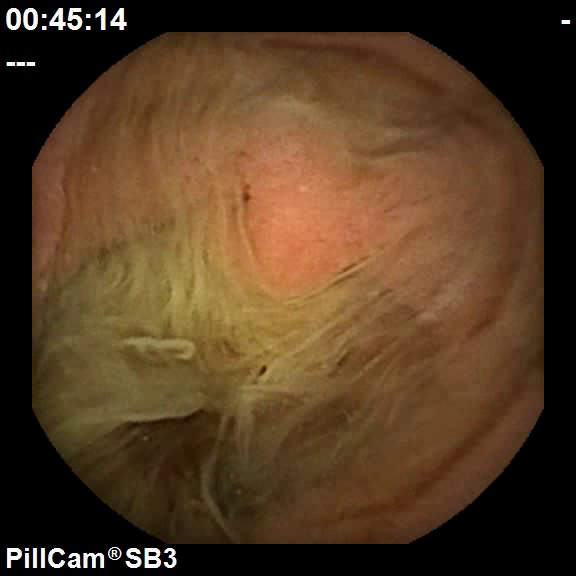}
     \end{subfigure}
    \hspace{-.750cm}
   \rule{1pt}{23pt}
    \hspace{-.750cm}
    \begin{subfigure}[b]{0.1\textwidth}
         \centering
         \includegraphics[trim=32 32 32 32,clip,scale=0.04]{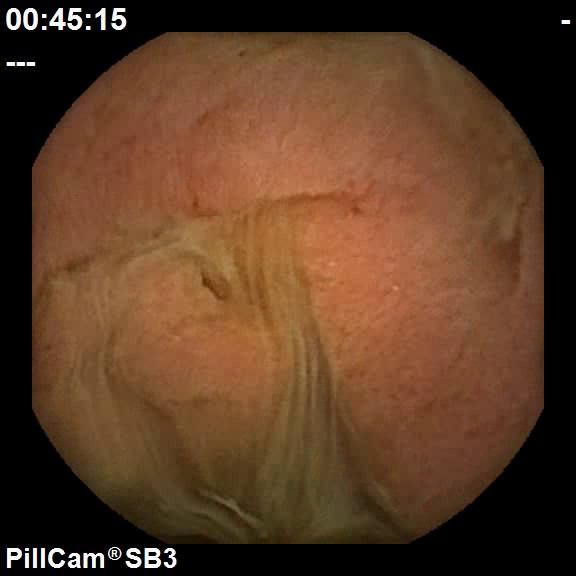}
     \end{subfigure}
     \hspace{-1.35cm}
      \begin{subfigure}[b]{0.1\textwidth}
         \centering
         \includegraphics[trim=32 32 32 32,clip,scale=0.04]{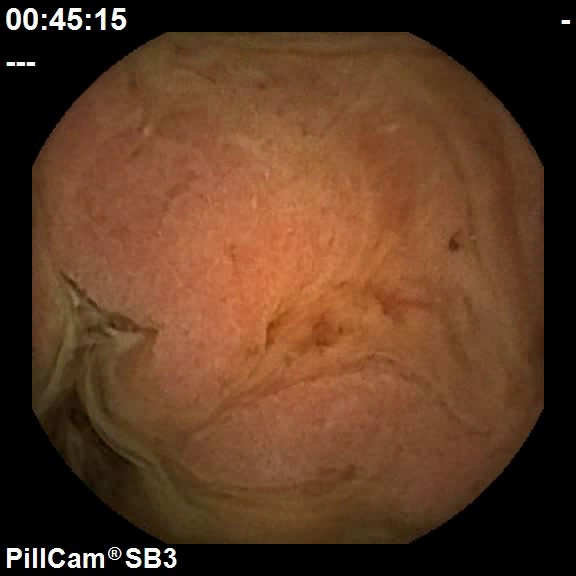}
     \end{subfigure}
     \hspace{-1.35cm}
      \begin{subfigure}[b]{0.1\textwidth}
         \centering
         \includegraphics[trim=32 32 32 32,clip,scale=0.04]{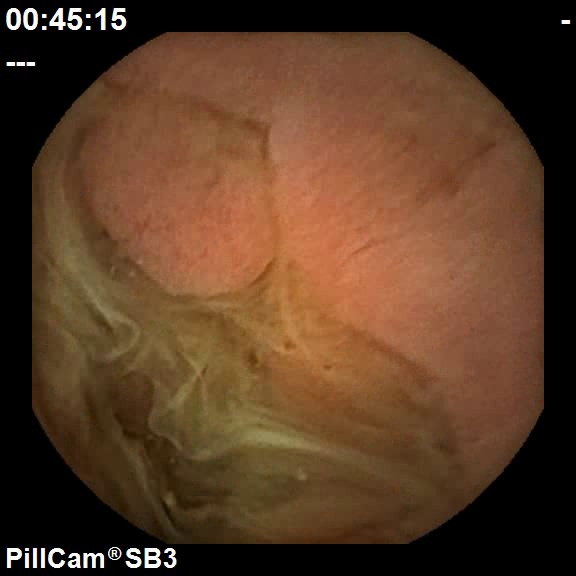}
     \end{subfigure}
    \hspace{-.750cm}
   \rule{1pt}{23pt}
    \hspace{-.750cm}
      \begin{subfigure}[b]{0.1\textwidth}
         \centering
         \includegraphics[trim=32 32 32 32,clip,scale=0.04]{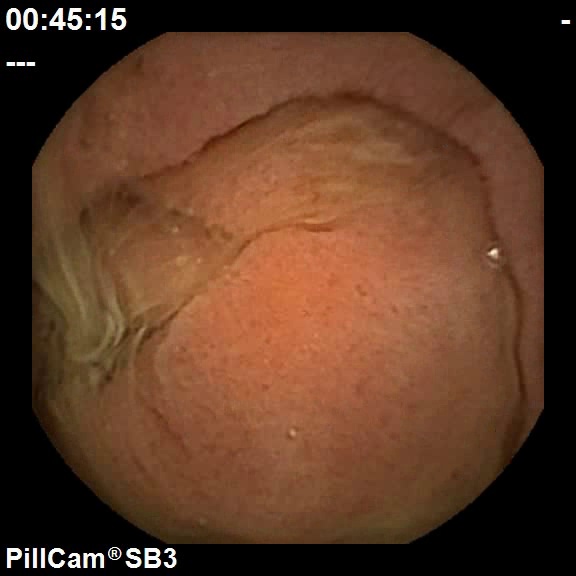}
     \end{subfigure}
    \caption{Visual Illustration of Detected Video Boundaries}
    \label{fig:sbds}
\end{figure}

Figure \ref{fig:sbds} shows the output of the boundary detection step.

For a long CE video with multiple diseases at different regions, we applied unsupervised temporal segmentation method using the PELT change point detection algorithm to split the videos features into homogeneous visual segments. Each segment is then considered a bag of normal and abnormal frames with some frames containing only normal frames. A video segment is considered abnormal if it contains at least one frame with an abnormality. Similar to a Multi-instance learning problem, an abnormal video segment contains a mix normal and abnormal frames \cite{zhou2004multi, foulds2010review, maron1998framework}.

\begin{figure*}
    \centering
    \includegraphics[width=0.9\textwidth, height=8.5cm]{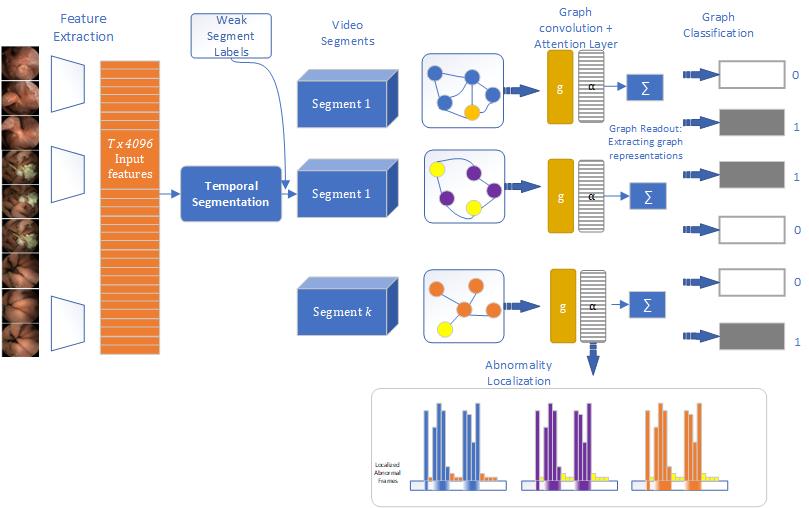}
    \caption{Weakly Supervised Abnormality Localization Model}
    \label{fig:loc_model}
\end{figure*}

Each unique patient's video $V^{n}$ is temporally segmented into $k$ video shots $ \{v_{i=0}^{n}, ..., v^{n}_{i=k}\}$ based on the visual temporal boundaries. While the long video $V^{n}$ can contain multiple diseases and therefore have multiple labels $ \{y_{1}, ... , y_{l}\}$, the result of the segmentation step allows us to only capture one or no abnormality within each short video segment. The illustration in shown in figure \ref{fig:loc_model}. First, we partition the videos into uniform segments $ \{v_{i=0}^{n}, ..., v^{n}_{i=k}\} $ such that not more than abnormality is present in each segment with no overlapping frames. We considered a disease agnostic framework with labels $y_{i} \in \{0, 1\}$ such that we only classify each segment as either abnormal or normal based on whether it contains at least an instance of an abnormal frame. This binary disease-agnostic framework will allow our model to generalize to any unseen category of abnormality in the future.

\subsection{Graph Representation and Classification}
We applied GraphSage convolution framework from \cite{Hamilton2017Jun}. The framework allows for inductive and transductive learning on large graphs. The model architecture is shown in figure \ref{fig:loc_model}. The input to the model is the extracted frame features for each video segment t x d where t is the length of the video segment (i.e. number of frames) and d is the dimension of the feature vectors.

During training, we only have access to weak labels for the video segments as shown in fig. \ref{fig:loc_model}. While we know there is a certain abnormality in the video, we do not have granular information of the frames where the disease is captured. Physicians also use the frequency of occurrence of a disease in multiple frames to determine its severity. A fully supervised node classification model will utilize labels pointing to the actual frame containing the disease as localizing the frame with the disease is important in helping the physician make quick and proper diagnosis. We consider a video segment as a bag of normal and abnormal frames i.e. given a video $v \in \mathrm{R}^{h \text{x} w \text{x} t}$ where $h$ and $w$ are the height and width of the frames and $t$ is the number of video frames in the segment. We consider $\mathcal{V} = \{f_{n},f_{a}\}$ where $f_{n}$ and $f_{a}$ are normal and abnormal frames respectively. A single CE video may contain multiple diseases or abnormalities making the abnormal class a combination of different abnormalities or diseases. This class agnostic model makes the model generalize to other new unseen diseases in the future.

Following the above, we define a graph $\mathcal{G} = \{\mathcal{V}, \mathcal{E}\}$ with nodes $\mathcal{V}$ representing the frames in the video and edges $\mathcal{E}$ representing the connections between the frames. Secondly, we denote a sub-graph for each video segment $g = \{ v, e\}$ and edges $e$ representing edges between the frames in the video segment. Recall that each sub-graph $v$ is a bag containing both normal and diseased frames occurring at different points in the video. Our goal are in two stages, First is the multi-instance graph classification where, for any video segment containing at least one abnormal frame, we predict abnormal label otherwise, we predict abnormal label. Next is the abnormality localization where we generate frame-level activation score the abnormal video segment. The goal of the video segment classification is to first learn a mapping of $\psi(\mathcal{G}) \rightarrow \{y_{i}\}_{i=0}^{1}$ to binary normal and abnormal video segment. In our case, we employed disease agnostic binary category so as to be able to generalize to any unseen videos of new patient. The GraphSage network, through the sequence of transformation, aggregation, attention and multi-instance classification learns to classify each video segment into binary category of normal and abnormal segment. Next is to use the parameters of the learned network and the sequence of linear transformation, aggregation and final attention layer to score the frames in the abnormal segment based on their contribution to the graph prediction. This step is our localization step which occurs only at test time.

\subsection{Graph Convolution Network}
The uniqueness of this work is the application to long videos where there may be more than action within the sequence. Secondly, this work advances other prior works through frame level localization as against localizing to temporal region or volume in the video. Lastly, rather that using a fixed length temporal video features as input to the GCNN network \cite{Rashid_2020_WACV, Zeng2019}, our video segment inputs have varying length based on detected shot boundary in the long video. This makes our framework a complete end-to-end localization framework which has not been previous addressed in literature. Such end-to-end automatic segmentation, classification and localization helps mitigate against any intersection and correlation between member frames in each video segment.

The graph convolution involves three main steps: 1) Neighborhood aggregation; 2) Node representation: which involve concatenation, linear transformation and non-linear activation steps; 3) graph read-out.

Steps (1) and (2) occur at each layer of the network, while step (3) occurs at the final layer of the network. For our model, we used two (2) graph convolution layer.

The input to the network are the frame feature sub-matrix where each frame-feature represents a node in the graph with the weighted edges computed as the similarity between the features. Each node is directly connected to every other nodes but the edge weights is set to be proportional to the level of similarity between the pair of nodes. Thus, each video graph is a \textit{complete graph}. Since all frames are images of different locations of the small bowel, we allow nodes to derive message from every other nodes in the graph. Secondly, the formulation allows feature similarity and dissimilarity to be incorporated into the parameter learning process. This similarity between edges, essentially, captures the topological relationship between the frames. GCNN explicitly ensures relationship between frames is put into consideration during both training and testing as it aggregates the neighbouring nodes into each node for every layer of the network. Each frame feature vector is transformed by a weighted average of all other neighbouring frames it is connected to with weights based on learned edge strengths. In our case, all the frames in the video is a neighbour but the edges are weighted by the similarity function. We applied cosine similarity defined in \ref{eqn:cosine_sim} as the similarity metric between pair of the frame feature vector. Frames without any similariy will have edge weight of zero - meaning no connection between them. Other similarity function such as KNN, correlation and euclidean distance were also experimented with and we compared the results across. The only problem with using a nearest neighbor relationship is having to set the number of neighbors $k$, which may not be optimal for the dataset.
\subsubsection{Feature Aggregation and Node Embedding}
\begin{figure}
    \centering
    \includegraphics[scale=0.45]{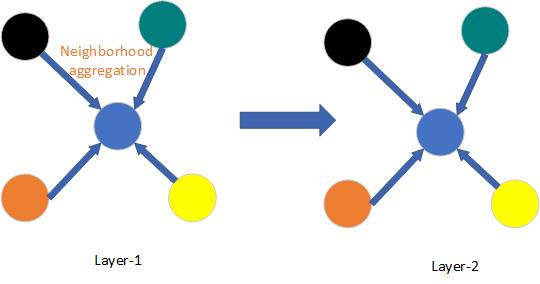}
    \caption{Neighborhood Aggregation}
    \label{fig:node_agg}
\end{figure}

Fig. \ref{fig:node_agg} shows a representation for he neighborhood feature aggregation. After the first layer, each node feature is a weighted average of all the neighboring node features.

Starting from the initial input features
\begin{equation}
    \boldsymbol{h}_{i}^{0} = \textbf{x}_{i}, \hspace{10pt} \forall i \in v;
\end{equation}
where $i$ represents a nodes (frames) and $v$ is the video segment.

the representation of the neighbors of node $i$ at layer $l+1$ is given as the weighted aggregation of all neighboring node $j$ features;

\begin{equation}\label{eqn:aggregation1}
    \boldsymbol{h}_{N(i)}^{l+1} = \text{AGGREGATE}\bigg(e_{ij} h_{j}^{l}, \hspace{5pt} \forall j \in N(i)\bigg);
\end{equation}
where $j$ represents neighboring node to node $i$.

Aggregation functions such mean, max-pool and LSTM can be applied. After experimenting with the different aggregator functions, LSTM outperformed the others and also more stable to train. We used the LSTM aggregation between each pair of the nodes. Eq \ref{eqn:aggregation1} becomes

\begin{equation}\label{eqn:saggregation2}
    \boldsymbol{h}_{N(i)}^{l+1} = LSTM\bigg(e_{ij} h_{j}^{l}, \hspace{5pt} \forall j \in N(i)\bigg);
\end{equation}
where $N(i)$ is the total number of neighbors of node $i$. For a complete graph, this will be one short of the total number of nodes in the graph. The LSTM aggregation steps are as follows in step eqn. \ref{eqn:lstm}:

\begin{align}\label{eqn:lstm}
    z_{l} = \sigma \bigg(W_{z} \cdot [h_{i}^{l}, e_{ij} h_{j}^{l}], \hspace{5pt} \forall j \in N(i)\bigg); \\
    r_{l} = \sigma \bigg(W_{r} \cdot [h_{i}^{l}, e_{ij} h_{j}^{l}], \hspace{5pt} \forall j \in N(i)\bigg); \\
    \Tilde{h}_{l} = tanh\bigg(W \cdot [r_{l} * h_{i}^{l}, e_{ij} h_{j}^{l}], \hspace{5pt} \forall j \in N(i)\bigg); \\
    \boldsymbol{h}_{N(i)}^{l+1} = (1-z_{l}) * h_{i}^{l} + z_{l} * \Tilde{h}_{l}
\end{align}

Next, we get the embedding for node $i$ by concatenating neighboring nodes representation $\boldsymbol{h}_{N(i)}^{l+1}$ with the previous layer embedding of node $i$ itself;

\begin{equation}\label{concatenation}
    \boldsymbol{h}_{i}^{l+1} = \sigma \bigg(\boldsymbol{W}^{k} \cdot \text{CONCAT}  \bigg({h}_{i}^{l},{h}_{N(i)}^{l+1} \bigg)\bigg)
\end{equation}
Eq. \ref{eqn:saggregation2} is the aggregation of the features from all connected neighboring nodes weighted by the edge similarity.
\subsubsection{Graph Attention and graph aggregation Layer}
After the final layer of the graph convolution operation, we applied an attention layer over the node embedding. The attention layer allows us to learn a parametric weighting of the nodes based on their importance to the graph classification. This allows the model to learn to place more weight on abnormal frames for the video segments with abnormality as also the most relevant frames for segments that are completely normal. We learn a representation of the entire GCN network at that last layer by aggregating features from all the nodes. Attention-based LSTM and GRU have been report effective in learning similar representation over sequences \cite{liu2017global}. However, GCNN model allows additional flexibility over a wide range of representation from mean to max-pooling over the nodes to the more complex LSTM aggregation at this layer. This final graph aggregation is called the graph readout layer.


\begin{equation}\label{readout}
    h_{g} = \frac{1}{N} \big( \sum_{i} \alpha_{i} h_{i} \big)
\end{equation}

where $h_{g}$ is the representation of the entire graph $g$ of the video segment $V_{k}$. Other readout operations include mean, summation and max-pool over the nodes embedding learnt across the layers of the network.

\subsubsection{Multi-Instance Graph Classification}
Once we aggregate the graph into a single feature vector, the final graph classification layer is a fully-connected layer that maps the graph embedding to the number of categories in our dataset before applying a sigmoid layer. We predict the binary label for each graph as follows:

\begin{equation}
    \hat{y}_{i=1}^{N} = \frac{1}{1 + e^{-h_{g}}}
\end{equation}
Where $N$ is the number of graphs and $h_{G}$ is the learned representation of $g$.
 
\begin{figure}
    \centering
    \includegraphics[scale=0.45]{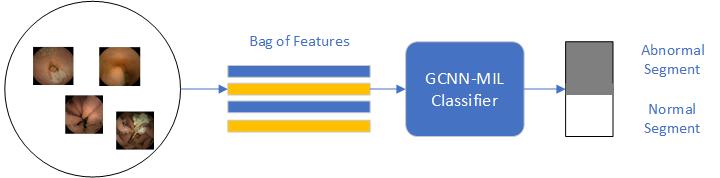}
    \caption{Multi-Instance Graph Classification}
    \label{fig:graph_mil}
\end{figure}

Fig. \ref{fig:graph_mil} shows the illustration of the multi-instance graph classifier.



\begin{equation}\label{eqn:cosine_sim}
    \boldsymbol{e}_{i,j} = \frac{\boldsymbol{x}_{i}^{T}\boldsymbol{x}_{j}}{||\boldsymbol{x}_{i}||_{2} \cdot ||\boldsymbol{x}_{j}||_{2}}
\end{equation}



\subsection{Graph Localization Network}
The graph localization network is the second step after training the parameters $W$ of $g$. The step occurs during testing, using the trained parameters $W$, we replaced the final graph readout function with a temporal pool layer to allow us generate a class activation map over the sequence of frames in the video. Our localization network generates ranking for each node in the graph. We sampled the temporal pool layer to identify nodes with the abnormality. Since abnormal regions in CE video is not necessarily contiguous, temporal pool over frames better captures the localization than temporal boundary regression \cite{Gao2017May}. Non-contiguity of abnormal regions is a unique property of CE videos which differentiates it from other video structured data.
\section{Experiments}\label{sec:experiments}
\subsection{Dataset Description}
Our dataset consist of nine (9) long VCE videos collected during real clinical endoscopy procedure under the supervision of expert gastroenterologist. All IRB requirements and approval processes were completed prior to analyzing the data. Since physicians are more interested in the small bowel region for the CE video examination, we focused our analysis on images of the small bowel region only. Each video was carefully annotated by two (2) endoscopy research scientist and verified by an expert gastroenterologist. We fine-tuned the pretrained feature extractor network on the first five (5) videos in our dataset and used it to extract features for all other videos. Since each video is unique to each patient, we ensured separation between videos that have been previously seen by the model were not part of the test videos. This helps mitigate patient bias.

The training video and the diseases captured in training data one is shown in table \ref{tab:train_data}

\begin{table}[htbp]
\caption{Training \& Test Video Data Description}
\begin{center}
\begin{tabular}{|c|c|c|}
\hline
 & \multicolumn{2}{c|}{\textbf{Video Content}} \\
\cline{2-3} 
\textbf{ Train Video} & \textbf{\textit{Nodes Count}} & \textbf{\textit{Abnormal Categories}} \\
\hline
Video - 1 & 13,177 &  Normal, Erythema, \\
& &  Outgrowth (Mass) \\
\hline
Video - 2 & 8,452 & Normal, Angioectasia, Diffuse bleeding,  \\
& & Erosion, Erythema, Ulcer \\
\hline
Video - 3 & 23,124 & Normal, Diffuse bleeding, Ulcer,  \\
& & Angioectasia, Outgrowth \\
\hline
Video - 4 & 12,303 & Normal, Angioectasia, Outgrowth \\
& &  Erythema, Erosion, Clot \\
\hline
Video - 5 & 29,236 & Normal, Bleeding, Ulcer \\
& &  Erythema \\
\hline
\textbf{Total} & \textbf{86,292} & \\
\hline
\hline
 & \multicolumn{2}{c|}{\textbf{Video Content}} \\
\cline{2-3} 
\textbf{ Test Video} & \textbf{\textit{Nodes Count}} & \textbf{\textit{Abnormal Categories}} \\
\hline
Video - 6 & 14,173 &  Normal, Ulcer, Angioectasia, \\
& & Erythema, Erosion \\
\hline
Video - 7 & 16,909 & Normal, Bleeding  \\
\hline
Video - 8 & 10,037 & Normal, bleeding, Angioectasia \\
\hline
Video - 9 & 19,104 & Normal, Bleeding, Ulcer \\
\hline
\textbf{Total} & \textbf{60,223} & \\
\hline
\end{tabular}
\label{tab:train_data}
\end{center}
\end{table}

In our proposed model (shown in Figure: \ref{fig:loc_model}), the score predicted for each frame corresponds to the node activation sequence for the frame. Rather than using the granular class of each abnormality shown in table \ref{tab:train_data}, we used a class-agnostic binary label for the graph classification. This allows the model to generalize to any unseen categories of abnormalities in future videos.

\subsection{Evaluation}
We evaluated our proposed framework in two folds. First, the performance of the multi-instance graph classification model was evaluated on new patients' test videos based on accuracy, sensitivity, specificity and f-score. Evaluation based on the intersection-over-union that has been employed in literature on localization does not directly apply on CE video since an abnormal temporal bound may not be contiguous. Instead, we employed the widely adopted evaluation framework on CE videos \cite{iakovidis2010reduction, tsevas2008automatic} - \textit{Coverage}. Which also is a measure of specificity of the model on the abnormal frames. The specificity of the abnormal classes is the most important criteria on which medical experts base the performance of machine learning models since this impacts the accuracy of their diagnosis. The coverage is defined as in equation \ref{eqn:coverage} which is the number of selected sample frames as a proportion of all abnormal frames in the segment. We aggregate this over the entire video to report our result.

\begin{equation}\label{eqn:coverage}
    \text{C} = \frac{\sum_{i}^{N_{ab}} c_{i}}{N_{ab}};
    c_{i} = \begin{cases} 1, & \text{Abnormal frame is selected} \\
            0, & \text{otherwise}
            \end{cases}
\end{equation}

where $N_{ab}$ is the count of video segments with abnormality. From \ref{eqn:coverage} The metric scores one (1) if at least one abnormal frame is selected and zero otherwise.

\subsubsection{Implementation}
Our entier model was implemented in Pytorch \cite{NEURIPS2019_9015} on NVIDIA RTX2080 GPU. We trained the GCNN using stochastic gradient descent optimization algorithm using cross entropy loss function with a batch size of 8 and learning rate of 0.001. The models were trained for a minimum of 100 epochs.

\section{Results and Discussion}\label{sec:results}
Table \ref{tab:graph_results} shows the result of the binary multi-instance graph classification task applied on four (4) different video data. The four (4) test videos are different from the training videos and have never been seen before by our model. This allows generalization of our model to new patients' videos. The total segments is the count of the abnormal and the normal video segments and the disease categories is the number of different diseases present in the complete video. Table \ref{tab:loc_results} shows the performance on the localization task. The result in both tables is the weighted average of the metrics computed over the binary classes which accounts for the class imbalance in the dataset.

\begin{table}[htbp]
\caption{Video Graph Classification Results}
\begin{center}
\begin{tabular}{|l|c|c|c|c|}
\hline
\textbf{Metrics} & \multicolumn{4}{c|}{\textbf{Test Video Data}} \\
\cline{2-5} 
 & \textbf{Video 6} & \textbf{Video 7} & \textbf{Video 8} & \textbf{Video 9} \\
 \hline
\textbf{Frames Count} (T) & 14,173 & 16,909 & 10,037 & 19,104 \\
\hline
\textbf{Total Segments} & 770 & 1,124 & 248 & 1,071 \\
\hline
\textbf{Disease Categories} & 5 & 2 & 3 & 3 \\
\hline
\hline
\textbf{Accuracy} & \textbf{0.899} & 0.848 & 0.560 & 0.859 \\
\hline
\textbf{Sensitivity} & \textbf{0.911} & 0.804 & 0.601 & 0.889\\
\hline
\textbf{Specificity} & \textbf{0.899} & 0.848 & 0.560 & 0.859 \\
\hline
\textbf{F-score} & \textbf{0.905} & 0.822 & 0.578 & 0.873 \\
\hline
\end{tabular}
\label{tab:graph_results}
\end{center}
\end{table}

On video-1, the model achieved classification accuracy of 89.9\% on 770 video segments with 5 different categories of diseases. The sensitivity, specificity and F-score are 91.1\%, 89.9\% and 90.5\% respectively. The best performance was recorded on video-1 indicating that the performance across patients are not equal and some patients' videos may be more challenging than others. With different number of classes across each of the videos, the result of the model reflects the realistic output when a new patient's video is shown to the model. Prior to administering the capsule endoscopy, patients are advised not to eat or consume any opaque liquid that could obstruct the visibility of the camera. Occlusion and other factors in the digestive tract varies across patients leading to difference in classification performance. On the segment classification task, the model performed least on video-3 with classification accuracy of 56.0\%; sensitivity of 60.1\%; specificity of 56.0\% and F-score of 57.8\%. The performance on the other two videos are better and closer to the performance on video-1. With the highest number of disease classes in video-1, the performance on video-2 makes it rather difficult to believe the number of different abnormalities present in the video may impact the performance of the multi-instance classifier. 

From table \ref{tab:loc_results}, for video-1 at k=1, by sampling a single (1) frame from each abnormal video segments the model is able to cover 92.5\% of all the abnormalities in the video. Similarly, by sampling the top-2 frames, the models covers 97.5\% of all abnormalities in the video. This, however, flattens after this point which may be attributed to a number of reasons. The number of activated high energy frames in the video segments is a proportion of the total number of frames that captures abnormality and the total length of the video segment. With very few (e.g. only 1) abnormal frames and very long video segment, it may difficult for the model to identify this single frame within the segment.  

The performance on the localization task does not exactly mirror the graph classification when looking across patients' videos. However, the trend is that the more the number of high energy frames selected, the higher the coverage that is obtained. While this may appear obvious, the performance varies across the videos with video-2,3 and 4 requiring a minimum of 9-high activation frames to achieve the same coverage obtained on video-1 with just 2-samples. 
High coverage means high true positive rate and indicates the model is able to accurately identify and rank frames leading to the output of the multi-instance graph classifier for abnormal graphs. Very high coverage will also allow the physician to only focus and examine the few selected localized frames as against having to review the entire video which would be much more time consuming. For example, in video-1, by selecting a sample frame from each abnormal video segment, physician will only have to review 40 frames to make their diagnosis as against the entire 14,173 of the small bowel region. On the other hand, a low coverage indicates high false positive (FP) leading to frames that do not contain any abnormality being selected as high energy frame. This will lead to increase sample frames that physician will have to review and analyse, thereby saving them less time and effort.

\begin{table}[htb]
\caption{Results of Abnormality Localization using Adaptive Temporal Pool Node Sampler}
\begin{center}
\begin{tabular}{|l|c|c|c|c|}
\hline
\textbf{Metrics/Data} & \multicolumn{4}{c|}{\textbf{Test Video Data}} \\
\cline{2-5}
 & \textbf{Video 6} & \textbf{Video 7} & \textbf{Video 8} & \textbf{Video 9} \\ 
 \hline
\textbf{Frames Count} (T) & 14,173 & 16,909 & 10,037 & 19,104 \\
\hline
\textbf{Abnormal Segments ($N_{ab}$)} & 40 & 137 & 57 & 69 \\
\hline
& \multicolumn{4}{c|}{\textbf{Coverage} ($C = \sum c_{i}/N_{ab}$)} \\
\hline
\textbf{k=1} & \textbf{0.925} & 0.467 & 0.667 & 0.391 \\
\hline
\textbf{k=2} & \textbf{0.975} & 0.664 & 0.772 & 0.464 \\
\hline
\textbf{k=3} & \textbf{0.975} & 0.752 & 0.825 & 0.638 \\
\hline
\textbf{k=5} & \textbf{0.975} & 0.883 & 0.825 & 0.797 \\
\hline
\textbf{k=7} & \textbf{0.975} & 0.956 & 0.877 & 0.913 \\
\hline
\textbf{k=9} & \textbf{0.975} & 0.912 & 0.912 & 0.928 \\
\hline
\end{tabular}
\label{tab:loc_results}
\end{center}
\end{table}

\begin{figure}
    \centering
    \includegraphics[trim=5 30 5 5, clip, scale=0.52]{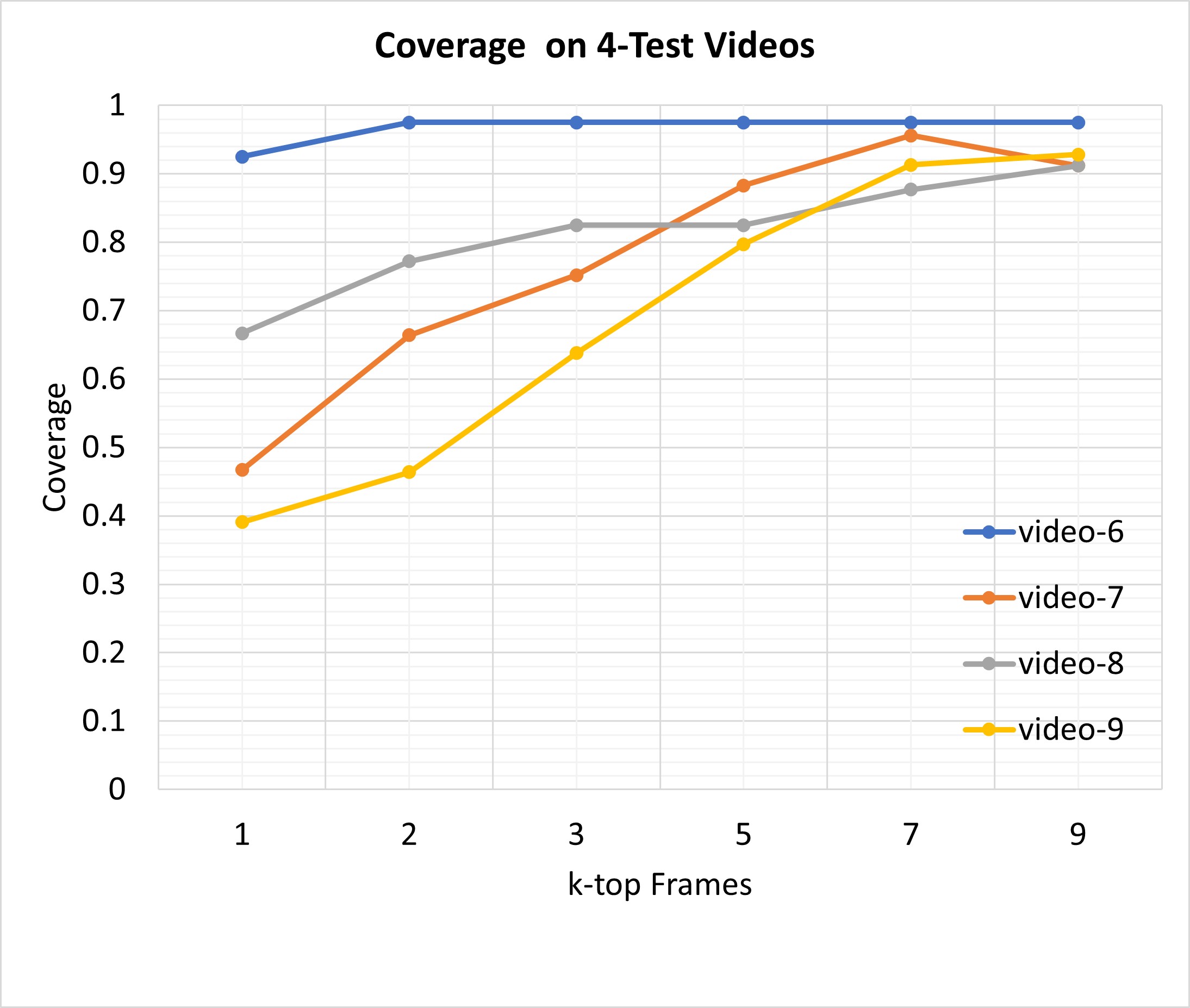}
    \caption{Performance on Abnormality Coverage}
    \label{fig:coverage}
\end{figure}
\section{Conclusion and Future Works}\label{sec:conclusion}
In this paper, we developed a novel end-to-end temporal abnormality localization for long wireless capsule endoscopy video using only weak video level annotation. We achieved the abnormality localization in three-steps, first is the long video temporal segmentation, then video segment classification before finally localizing to the high energy frames within each segment. In the classification step, our model learns to identify abnormal video segments from the aggregated embedding feature vectors using multi-instance learning framework. The localization step involves leveraging the representation of the graph to generate the high energy frames from each abnormal video segments. The end-to-end system involves, first applying an unsupervised temporal segmentation technique to partition the long WCE video into short, homogeneous segments. Thereafter, we trained a Graph Convolution Neural Network (GCNN) on each video segment to classify them into binary categories. We consider each video segment as a graph and the frame features as the nodes in the graph. We learnt a representation of the video segments using a 2-layer graph convolution. We applied attention layer on the nodes embedding before aggregating the node features at the final layer to generate the graph representation. The final layer is a multi-instance graph classifier that classifies the video segment feature vector into binary class-agnostic categories. Leveraging the parameters of the trained GCNN model, we replaced the final classifier with a temporal pool layer to select the most activated frames within the video segment which represents the highest energy elements of the graph.  
Similar to a video summarization model, the approach proposed in this paper for CE video abnormality localization allows physicians and gastroenterologist to quickly focus and review identified abnormal frames that captures abnormal lesion or diseases in more detail as against having to wade through the entire long CE video with thousands of redundant normal frames.

For our future works, we will consider a full graph classification using multi-label, multi-instance learning framework obviating the need for our temporal segmentation step. Since a long WCE video will typically contain multiple diseases or abnormalities, similar to multiple actions within a long video, a multi-label framework will allow us to simultaneously learn a representation and also classify the full graph into the multiple categories before employing the classifier to localize the frames with the highest activation in the video.


\bibliographystyle{IEEEtran} 
\bibliography{main}

\end{document}